\documentclass[review]{elsarticle}
\graphicspath{{./}} 

\usepackage{float}               
\usepackage{apalike}             
\usepackage{multirow, makecell}  
\usepackage{booktabs}            
\usepackage{geometry}            
\usepackage{amsmath, amssymb}    
\usepackage{tabularx}            
\usepackage{graphicx}            
\usepackage{placeins}            
\usepackage{natbib}              
\usepackage{xcolor}              
\usepackage{subfigure}   
\usepackage{adjustbox}   
\restylefloat{figure}
\floatstyle{plaintop} 
\restylefloat{table}

\journal{Expert\_Systems\_with\_Applications}

\biboptions{authoryear}

\usepackage{hyperref}
\hypersetup{
	pdfauthor={First A. Author et al.},  
	pdftitle={SFAM: Preparation of papers for Computational Visual Media},
	pdfencoding=unicode  
}
\makeatletter
\ifdefined\@corref
\renewcommand{\@corref}[1]{}
\fi
\ifdefined\cnotenum
\renewcommand{\cnotenum}[1]{}
\else
\newcommand{\cnotenum}[1]{}
\fi
\makeatother

\begin{document}
	
	\begin{frontmatter}
		\title{S$^2$F-Net: A Robust Spatial-Spectral Fusion Framework for Cross-Model AIGC Detection}
		
		\author[aff1]{Xiangyu Hu\textsuperscript{$\dagger$}}  
		\ead{2529562337@qq.com}
		\author[aff1]{Yicheng Hong\textsuperscript{$\dagger$}}  
		\ead{15999928720@163.com}
		\author[aff2]{Hongcheng zheng}  
		\ead{zhc@e.gzhu.edu.cn}
		\author[aff1]{Wenjun zeng}  
		\ead{3053751954@qq.com}
		\author[aff1]{Bingyao Liu\textsuperscript{$*$}}  
		\ead{authorA@email.cn}  
		
		\address[aff1]{School of Computer Science and Engineering, Guangdong Ocean University, YangJiang, 529500, China}
		\address[aff2]{School of Artificial Intelligence, Guangzhou University, Guangzhou, 510623, China}
		
		\cortext[cor1]{$*$ Corresponding author: Bingyao Liu. Tel: +XX-XXXX-XXXXXXX; E-mail: authorA@email.cn.}
		\cortext[co1]{$\dagger$ These authors contributed equally to this work.}
		\begin{abstract}
			The rapid development of generative models has imposed an urgent demand for detection schemes with strong generalization capabilities. However, existing detection methods generally suffer from overfitting to specific source models, leading to significant performance degradation when confronted with unseen generative architectures. To address these challenges, this paper proposes a cross-model detection framework called S$^2$F-Net, whose core lies in exploring and leveraging the inherent spectral discrepancies between real and synthetic textures. Considering that upsampling operations leave unique and distinguishable frequency fingerprints in both texture-poor and texture-rich regions, we focus our research on the detection of frequency-domain artifacts, aiming to fundamentally improve the generalization performance of the model. Specifically, we introduce a learnable frequency attention module that adaptively weights and enhances discriminative frequency bands by synergizing spatial texture analysis and spectral dependencies.On the AIGCDetectBenchmark, which includes 17 categories of generative models, S$^2$F-Net achieves a detection accuracy of 90.49\%, significantly outperforming various existing baseline methods in cross-domain detection scenarios.
		\end{abstract}
		
		
	\end{frontmatter}
	
	\clearpage 
	\twocolumn 
	
	\section{Introduction}
	With the rapid development of generative artificial intelligence, images synthesized by algorithms such as Generative Adversarial Networks (GANs)\cite{2014arXiv1406.2661G} and diffusion models have achieved extremely high visual fidelity. The iterative evolution from Variational Autoencoders (VAEs)\cite{2013arXiv1312.6114K} and GANs to diffusion models has driven significant breakthroughs in synthetic images regarding visual quality, semantic complexity, and generation efficiency. Notably, after 2020, diffusion models gradually replaced GANs as the mainstream paradigm due to their superior training stability and high-fidelity output. However, the popularization of commercial generative AI has significantly lowered the threshold for generating fake images, leading to the severe challenge of the proliferation of false content. Consequently, developing robust AI-generated image detection methods is an urgent imperative.
	
	Current methodologies for generative image detection have converged into four core paradigms: spatial domain learning \cite{2013arXiv1312.6114K,2019arXiv191211035W,Liu_2020_CVPR,2023arXiv231210461T,2023arXiv231112397Z}, frequency domain analysis \cite{10.5555/3524938.3525242,2024arXiv240307240T,Qin2020FcaNetFC}, physical or reconstruction consistency \cite{2023arXiv230309295W,2024arXiv240317465L,2024arXiv241207140C}, and noise residual extraction \cite{10203908,10.1007/978-3-031-19781-9_6}. Spatial methods (e.g., CNNSpot \cite{Wang_2020_CVPR}, PatchCraft \cite{2023arXiv231112397Z}) scrutinize pixel-level statistics to identify texture distortions or spatial inconsistencies. In contrast, frequency domain approaches (e.g., FreDect \cite{pmlr-v119-frank20a}, FreqNet \cite{Tan2024FrequencyAwareDD}) utilize spectral transforms like FFT or DCT to isolate periodic artifacts unique to synthetic imagery. Beyond primary signals, higher-order pathways have emerged: reconstruction-based methods like DIRE \cite{2023arXiv230309295W} leverage the discrepancy in diffusion-based denoising errors, while noise-based techniques like LNP \cite{10.1007/978-3-031-19781-9_6} employ high-pass filtering to isolate forgery traces. Despite these advancements, mainstream models often suffer from poor generalization when encountering unknown sources. This decline stems from an over-reliance on semantic-level features or model-specific artifacts, failing to capture the universal "fingerprint" features inherent across diverse generative mechanisms. This limitation underscores the need for more robust, model-agnostic detection frameworks.
	
\begin{figure*}[h!t]
		\centering
		\includegraphics[width=17cm]{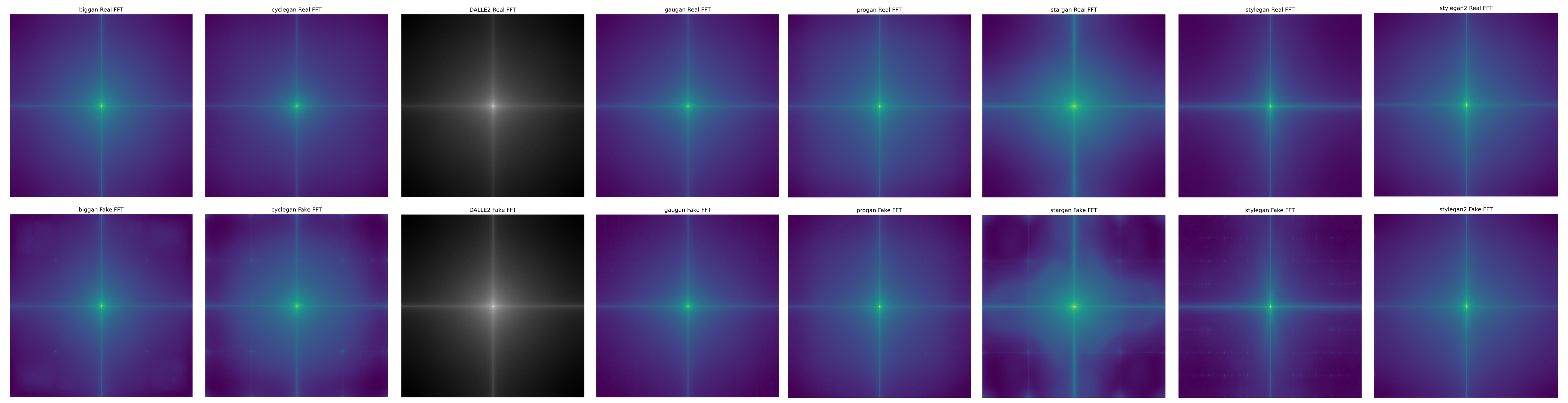}  
		
		
		\caption{Energy Comparison after Fourier Centralization}
		\label{FFT16}
	\end{figure*}

	Our research is grounded in a fundamental physical observation: \textit{while generative models can accurately simulate real images at the semantic level, their outputs inevitably bear synthesis traces due to inherent defects arising from cumulative upsampling operations within the generation pipeline.} Specifically, operations such as transposed convolution or interpolation disrupt the intrinsic inter-pixel statistical regularities of natural images, introducing periodic grid-like artifacts that manifest as distinct high-frequency anomalies in the frequency domain (as shown in Figure \ref{FFT16}). 
	
	\begin{figure*}[h!t]
		\centering
		\includegraphics[width=17cm]{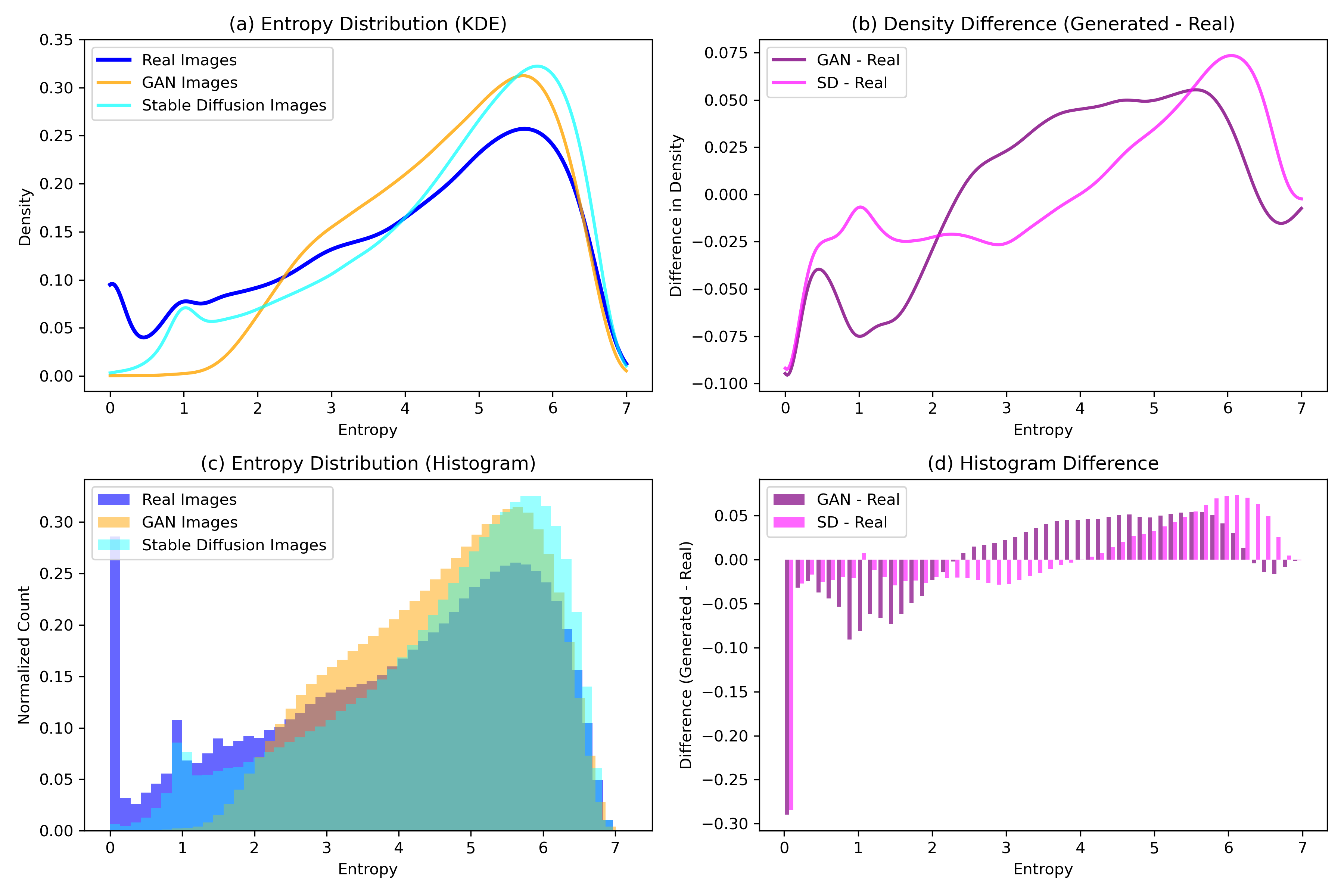}  
		
		
		\caption{Entropy Distribution Contrast (a) and (c) , Density Difference (b) and Histogram Difference (d) illustrates the differences in entropy
			region distribution between images generated by
			GANs and real images \& images generated by Stable Diffusion and real images.}
		\label{fig2}
	\end{figure*}
	
	Furthermore, in accordance with the F-principle of deep learning, generative models prioritize fitting low-frequency components during early training before progressively approximating high-frequency details. However, the fidelity of this high-frequency fitting process is significantly constrained by image entropy; in texture-rich (high-entropy) regions, generative models struggle to accurately reproduce the complex edge and texture distributions of real images. Consequently, the relative discrepancy between generated and real images is most prominent and stable in the high-frequency bands of high-entropy regions. As illustrated in Figure \ref{fig2}, compared to real images, GAN and Stable Diffusion models exhibit significant distributional fluctuations in high-entropy regions, while displaying abnormal smoothness in low-entropy regions. This distributional divergence further confirms that high-entropy regions serve as the critical locus for capturing forgery fingerprints.

	To address these challenges, we propose S$^2$F-Net, a detection framework that integrates spatial texture analysis with a frequency-domain attention mechanism. We first employ a "Smash \& Reconstruction" strategy \cite{2023arXiv231112397Z} to suppress global semantic interference, explicitly redirecting the model's focus toward local texture anomalies. Central to our innovation is the Learnable Frequency Attention (LFA) module. Departing from traditional methods that rely on fixed band-pass filters, LFA imposes distance-initialized learnable masks independently on the spectrograms of texture-rich and texture-poor branches to precisely isolate critical frequency bands. Subsequently, leveraging a group-wise weighting mechanism, the module computes the spectral energy within these adaptive bands and projects this energy into channel-wise attention weights. This architecture empowers the model to dynamically recalibrate spatial features based on spectral significance—specifically enhancing high-frequency artifacts in texture-rich regions and low-frequency anomalies in texture-poor regions—thereby effectively capturing universal cross-model fingerprints.
	
	The main contributions of this paper can be summarized as follows:
	\begin{enumerate}
		\item \textbf{Framework: Dual-Domain Disentanglement Network (S$^2$F-Net)}: We propose S$^2$F-Net, a novel dual-stream framework that strategically integrates spatial semantic stripping with adaptive frequency learning to decouple content from model-agnostic fingerprints, significantly enhancing generalization across unseen generative sources.
		\item \textbf{Module: Learnable Frequency Attention (LFA)}: We introduce a learnable frequency attention mechanism utilizing distance-initialized masks and group-wise weighting to precisely isolate and adaptively enhance discriminative frequency bands, transcending the limitations of fixed-filter approaches.
	\end{enumerate}

	\section{Related Work}
	Image generation technology is a core direction of both academic research and industrial implementation in the field of computer vision. Its primary objective is to approximate the distribution of real-world images from random noise or text descriptions through model learning, which has mainly undergone the evolution of two major technical paradigms: Generative Adversarial Networks (GANs) and diffusion models.
	Image generation technology aims to approximate the distribution of real-world data through model learning, and has undergone a paradigm shift from Generative Adversarial Networks (GANs) to Diffusion Models.
	
	\subsubsection{GANs Series Models}
	The Generative Adversarial Network (GAN) framework proposed by Goodfellow et al. \cite{2014arXiv1406.2661G} established the mainstream generation paradigm centered on the adversarial game between a generator and a discriminator. To address issues such as training instability and uncontrollable generation in the original GAN, subsequent research has continuously iterated and optimized the architecture design and training strategies. For instance, the Conditional Generative Adversarial Network (CGAN) proposed by Mirza et al. \cite{2014arXiv1411.1784M} successfully achieved directed control over generated content by introducing conditional variables such as class labels. 
	
	In terms of improving generation quality and resolution, the Progressive Generative Adversarial Network (ProGAN) developed by Karras et al. \cite{2017arXiv171010196K} adopted a layer-by-layer progressive training strategy from low to high resolution, significantly enhancing image details; while the BigGAN proposed by Brock et al. \cite{brock2018large} combined large-batch training with deep residual networks to achieve high-fidelity generation at $256 \times 256$ resolution on ImageNet. Furthermore, breakthroughs have been made in scenario-specific optimization: for example, StarGAN proposed by Choi et al. \cite{Choi_2018_CVPR} supports multi-domain style transfer with only a single model, and StyleGAN proposed by Karras et al. \cite{Karras_2019_CVPR} has become a benchmark model in the field of face synthesis through disentangled facial attribute control. 
	
	However, despite the great success of GANs, their inherent training stability issues such as Mode Collapse and vanishing gradients have limited their further development on more complex data distributions.
	
	\subsubsection{Diffusion Models}
	Inspired by non-equilibrium thermodynamics, Sohl-Dickstein et al. (2015) \cite{pmlr-v37-sohl-dickstein15} first proposed the diffusion generation framework, defining a generation pathway through "forward noise addition - reverse denoising". Although early models were slightly inferior in generation quality, the Denoising Diffusion Probabilistic Model (DDPM) proposed by Ho et al. in 2020 \cite{nichol2021improved} optimized the reverse denoising process by simplifying the training objective, significantly improving generation quality and establishing the mainstream status of diffusion models \cite{2022arXiv220900796Y}.
	
	Subsequently, to enhance semantic control, Nichol et al. (2021) \cite{2021arXiv211210741N} introduced the Classifier-free Guidance mechanism in GLIDE, which greatly improved the consistency between text and images. On this basis, researchers have further explored the optimization of efficiency and architecture: the Vector Quantized Diffusion Model (VQDM) developed by Gu et al. (2022) \cite{2021arXiv211114822G} integrated vector quantization technology to mitigate the loss of global contextual information, while the Adversarial Diffusion Model (ADM) created by Dhariwal et al. (2022) \cite{NEURIPS2021_49ad23d1} for the first time surpassed GANs in generation fidelity by introducing attention mechanisms and architectural improvements, thus greatly expanding the application boundaries of diffusion models.
	
	\subsection{Related Detection Methods}
	With the rapid evolution of deep generative models, detection paradigms for AI-generated Content (AIGC) images are primarily dedicated to exploring the intrinsic disparities between generated and real images across distinct feature domains. Existing detection approaches can be categorized into the following five core technical paths. The key research progress is summarized as follows:
	
	\begin{enumerate}
		\item \textbf{Spatial Domain Learning} \\
		\quad Spatial domain methods focus on leveraging Deep Convolutional Neural Networks (CNNs) to capture statistical anomalies directly from pixel-level data. Since generative models struggle to perfectly replicate the complex texture distributions of natural images, Wang et al. proposed CNNSpot \cite{Wang_2020_CVPR}, which employs data augmentation strategies to capture generator-specific artifact patterns. Similarly, Gram-Net introduced by Liu et al. \cite{Liu_2020_CVPR} detects cross-GAN models by exploiting differences in global texture statistics. To eliminate the interference of global semantic content on texture analysis, PatchCraft proposed by Zhong et al. \cite{2023arXiv231112397Z} designed a \textbf{Smash \& Reconstruction} strategy that forces the model to focus on local pixel correlations. Furthermore, NPR proposed by Tan et al. \cite{2023arXiv231210461T} characterizes forgery features by modeling relational traces between neighboring pixels.
		
		\item \textbf{Frequency Domain Analysis} \\
		\quad Upsampling operations in generative models often introduce periodic grid artifacts in the frequency domain, which serve as a critical physical cue for distinguishing real images from fake ones. Frank et al. \cite{pmlr-v119-frank20a} leveraged the Discrete Cosine Transform (DCT) to identify anomalous high-frequency energy decay patterns in generated images, and further proposed FreDect to extract frequency-domain fingerprints, which effectively mitigates spatial content interference. FreqNet proposed by Tan et al. \cite{2024arXiv240307240T} aims to deeply mine subtle high-frequency anomalies. To utilize frequency-domain information more efficiently, FCANet proposed by Qin et al. \cite{Qin2020FcaNetFC} enhances feature representation via a multi-spectral channel attention mechanism. However, such methods typically rely on fixed DCT bases or manual frequency partitioning, limiting their adaptability to dynamically varying artifact distributions across different generative models.
		
		\item \textbf{Physical Property Consistency Checking} \\
		\quad Methods in this category discriminate real and fake images based on the inverse process of generative models or physical laws. DIRE proposed by Wang et al. \cite{2023arXiv230309295W} leverages the inverse denoising and reconstruction mechanism of pre-trained diffusion models, demonstrating that real images typically exhibit larger reconstruction errors compared to generated ones. LaRE proposed by Luo et al. \cite{2024arXiv240317465L} shifts error calculation to the latent space to enhance discriminative power. Recently, Chu et al. observed that diffusion models struggle to accurately reconstruct mid-frequency information of real images, leading to the proposal of FIRE \cite{2024arXiv241207140C}, a detection method based on frequency-guided reconstruction errors.
		
		\item \textbf{Noise Residual Extraction} \\
		\quad To eliminate the interference of semantic content, researchers have attempted to extract more essential forensic traces from the noise or gradient domains. LGrad proposed by Tan et al. \cite{10203908} utilizes gradient maps of pre-trained CNNs to filter out semantic redundancy, thereby highlighting structural gradient anomalies introduced by generators. LNP proposed by Liu et al. \cite{10.1007/978-3-031-19781-9_6} extracts image noise patterns using denoising models; by analyzing the statistical properties of residual noise in both frequency and spatial domains, it precisely captures subtle noise artifacts remaining in high-fidelity generated images.
		
		\item \textbf{Multi-modal Detection} \\
		\quad With the proliferation of vision-language pre-trained models, leveraging multi-modal features to enhance detection generalization has emerged as a new trend. Yousif et al. utilized CLIP-ViT to extract features, achieving remarkable zero-shot detection capabilities \cite{2023JFM...957A...6Y}. UnivFD proposed by Ojha et al. \cite{2023arXiv230210174O} mines cross-modal universal fingerprints via lightweight adapters. To further enhance interpretability, SIDA proposed by Huang et al. \cite{2024arXiv241204292H} is based on Large Multimodal Models (LMMs); it unifies forgery detection, tampered region localization, and natural language explanation through \texttt{<DET> and <SEG>} tokens, driving forensic technology toward semantic interpretability.
	\end{enumerate}
	
	\section{Our work}
	
	\subsection{Design Motivation}
	The core of image generation lies in the data-driven paradigm of fitting the high-dimensional probability distribution $p_{data}(x)$ of real images. Although generative models minimize divergence metrics (e.g., KL or Wasserstein distance) to approximate the model distribution $p_{\theta}(x)$ to $p_{data}(x)$, high-resolution synthesis remains hindered by a critical technical bottleneck: the inherent defects of upsampling operations. During the mapping from low-dimensional latent spaces to high-dimensional pixel spaces, upsampling fails to accurately recover the high-frequency details and local pixel statistical correlations of natural images, inevitably introducing spectral anomalies that deviate from natural distributions.
	
	Mechanistically, upsampling requires expanding pixel information by a factor of $s^2-1$, yet low-resolution feature maps primarily encode global semantics, leaving high-frequency details to be synthesized via model inference. The overlapping computation in transposed convolutions triggers a weight accumulation effect, which disrupts natural statistical regularities, while interpolation operations tend to over-smooth edges. As highlighted by Durall et al. \cite{10.1007/978-3-030-58574-7_7}, upsampling amplifies distributional biases during training, causing anomalous frequency-domain patterns—such as phase misalignment and spectral loss—to manifest in synthetic outputs. These artifacts, which conflict with natural physical laws, serve as the fundamental evidence for AIGC detection.
	
	Existing AIGC detection methodologies primarily follow two paradigms: spatial-domain learning and frequency-domain analysis. Spatial-domain methods utilize pixel inputs to extract distributional discrepancies via deep classifiers (e.g., ResNet\cite{He2015DeepRL}, EfficientNet\cite{pmlr-v97-tan19a}). Conversely, frequency-domain methods employ transforms like FFT or Discrete Wavelet Transform (DWT) to capture structural artifacts, such as "checkerboard" patterns and high-frequency energy loss, induced by upsampling. However, single-domain frameworks exhibit significant performance bottlenecks when generalizing across heterogeneous generative architectures.
	
\begin{figure}[!htbp] 
		\centering
		\includegraphics[width=\linewidth]{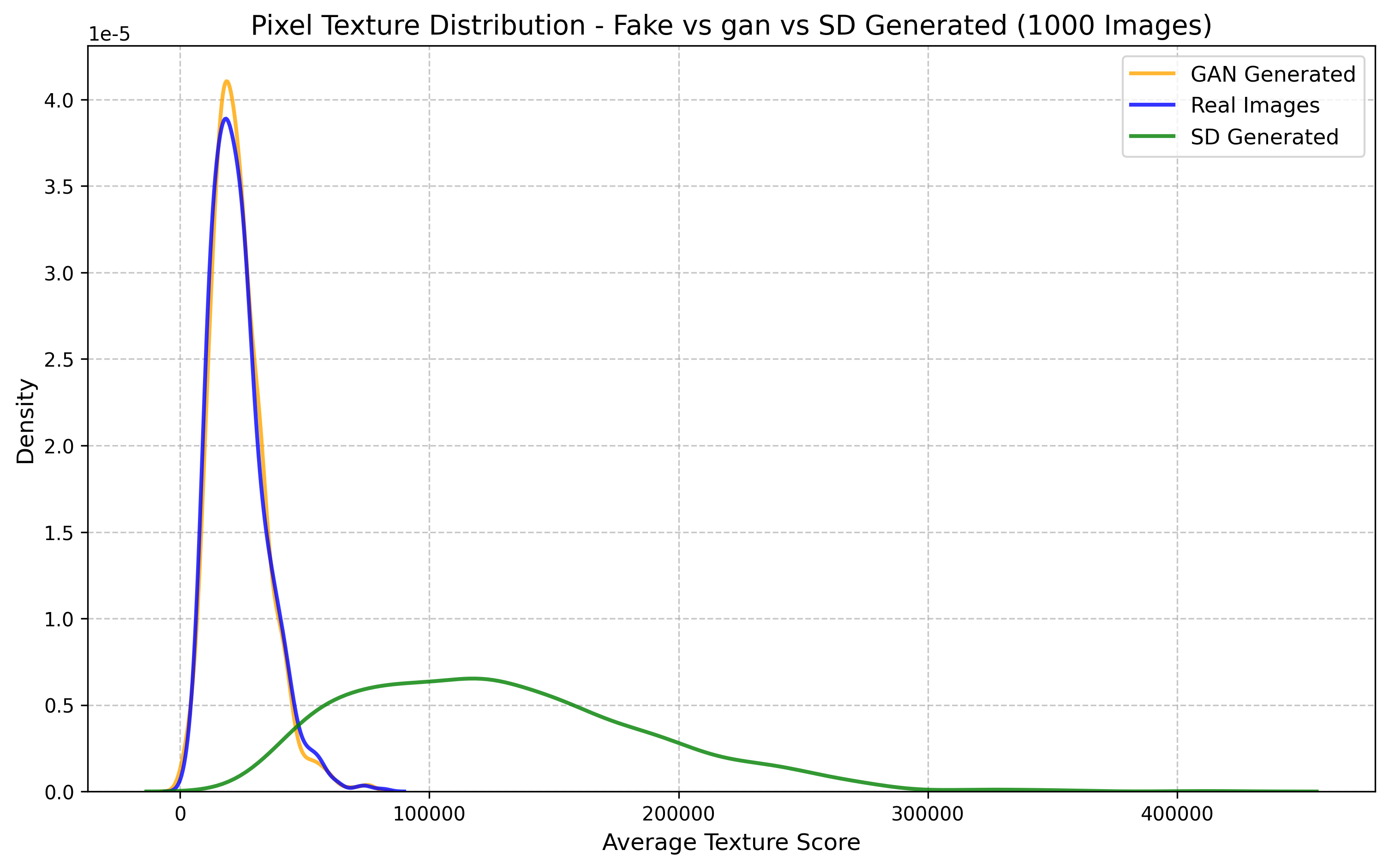}  
		
		\caption{Kernel Density Estimation (KDE) of texture richness distribution between real images and synthetic images from advanced generative models}
		\label{fig3}
	\end{figure}

	Spatial-domain methods rely heavily on global semantics and surface textures, struggling to decouple model-agnostic forgery features. As illustrated by the Kernel Density Estimation (KDE) in Figure \ref{fig3}, while the spatial texture richness of real images and GAN-generated samples shows partial overlap, the distribution of Diffusion models diverges drastically. This indicates that simplified texture complexity analysis is insufficient for establishing a robust decision boundary between advanced synthetic images and real content, leading to severe generalization decay on unseen sources.
	
\begin{figure*}[!htbp]  
		\centering
		\includegraphics[width=\textwidth]{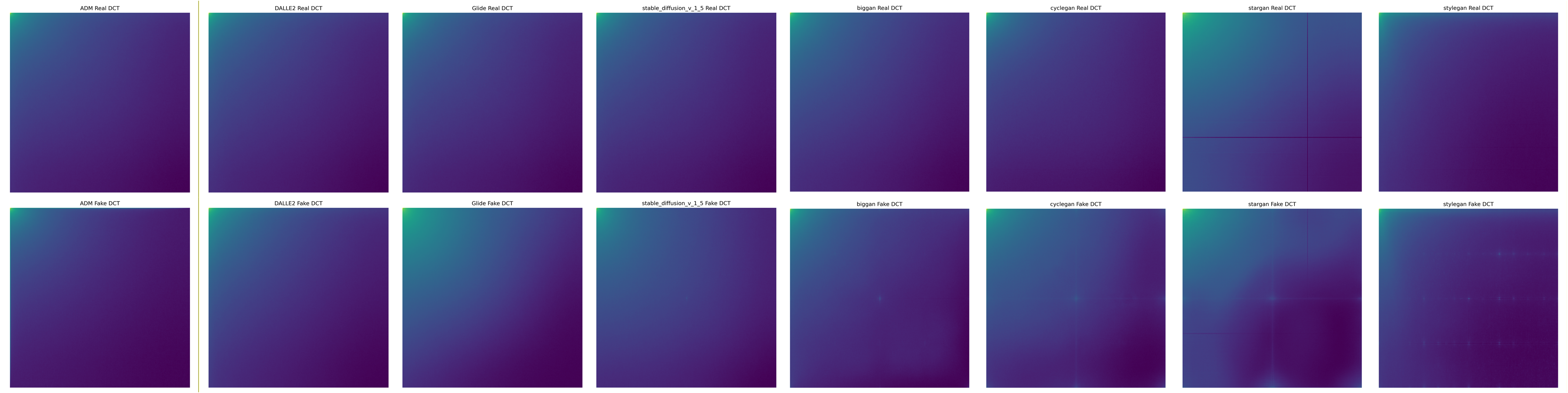}  
		
		\caption{Average Discrete Cosine Transform (DCT) heatmap comparison between GAN and Diffusion-generated images}
		\label{fig1}
	\end{figure*}

	Furthermore, traditional frequency-domain approaches are largely anchored to the fixed spectral artifacts characteristic of early GANs. Nevertheless, next-generation models have effectively suppressed these structured traces through refined upsampling strategies and sophisticated noise modeling. As evidenced by the average Discrete Cosine Transform (DCT) heatmaps in Figure \ref{fig1}, Diffusion models do not exhibit the anomalous high-frequency energy clusters typical of GANs. This "spectral smoothing" trend directly compromises the efficacy of detectors relying on fixed frequency fingerprints. These limitations collectively indicate that single-domain feature representation is no longer sufficient to address the increasingly complex AIGC detection tasks.
	
\begin{figure*}[h!t]
		\centering
		\includegraphics[width=17cm]{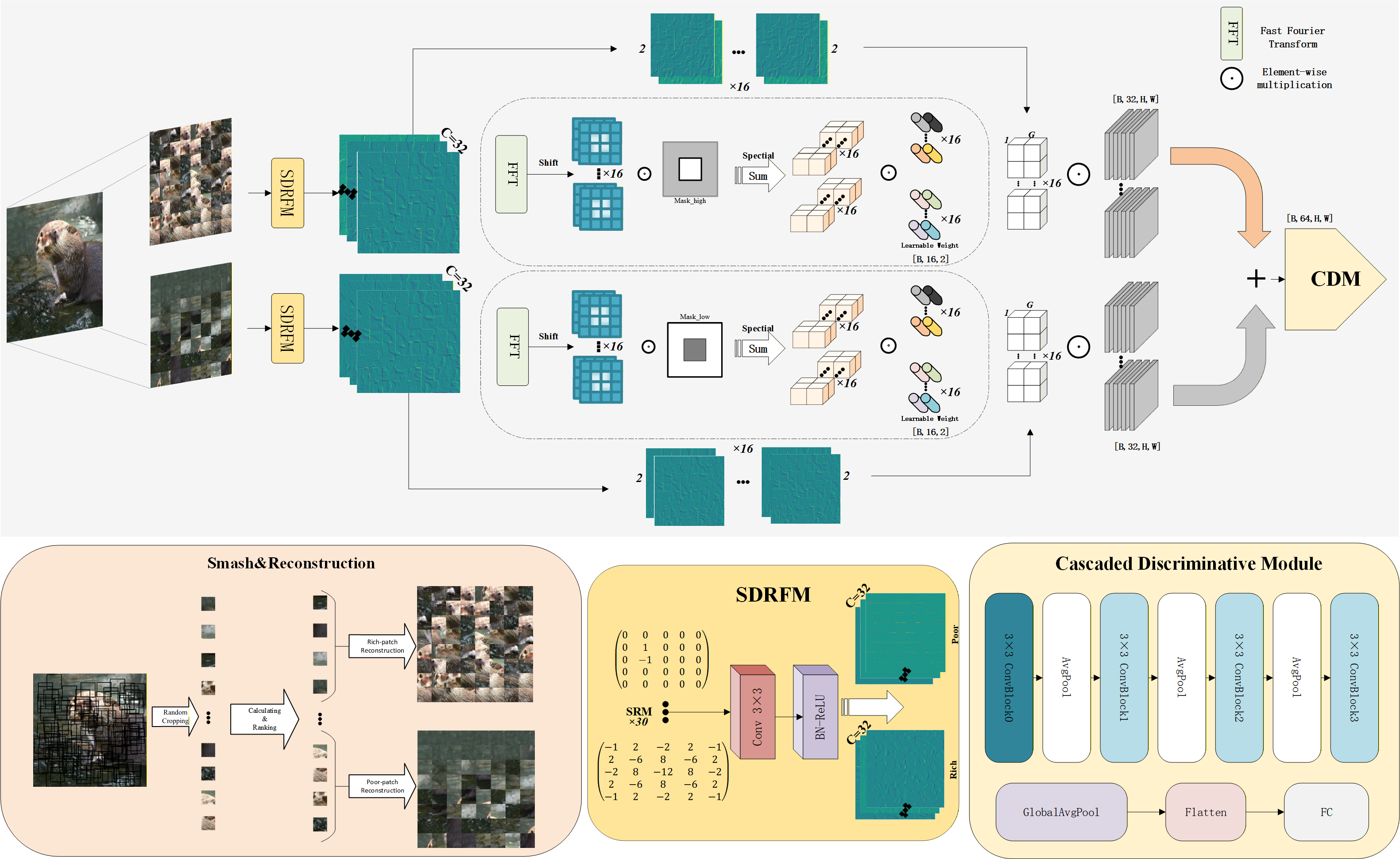}  
		
		
		\caption{S$^2$F-Net's overall architecture: integration of spatial feature extraction and learnable frequency attention mechanism}
		\label{fig4}
	\end{figure*}

	Given the complementarity between spatial textures and spectral energy balances, a dual-domain collaborative framework is essential to overcome the bottlenecks of generalization and robustness. Therefore, we propose S$^2$F-Net, a detection model integrating spatial feature extraction with a learnable frequency attention mechanism (overall architecture shown in Figure \ref{fig4}). The core of this framework consists of three functional modules: a spatial-domain feature extraction module, a frequency-domain attention weighting module, and a cascaded discrimination module.
	
	\subsection{Spatial Domain Feature Extraction: Semantic Smash and Residual Learning}
	To fundamentally suppress the interference of global semantic content on the detector, this module references the "Smash \& Reconstruction" strategy proposed by Zhong et al. \cite{2023arXiv231112397Z}, and combines it with lightweight residual mapping to construct a semantic-agnostic feature extraction pipeline.
	
	\subsubsection{Semantic Smash \& Physical Reconstruction}
	First, to physically disrupt semantic continuity at the pixel level, we utilize this strategy to reorganize the input image $I \in \mathbb{R}^{H_{in} \times W_{in} \times C}$. We perform random non-overlapping cropping to generate $N$ patches of size $M \times M$, where the cropping starting coordinates $(x_0, y_0)$ satisfy a uniform distribution:
	\[
	(x_0, y_0) \sim U(0, H_{in} - M) \times U(0, W_{in} - M)
	\]
	This yields a position-invariant set of image patches $\mathcal{P} = \{P_1, ..., P_N\}$. To quantify the statistical texture complexity of each patch, we calculate its four-directional pixel difference metric $l_{\text{div}}$:
	\[
	\begin{aligned}
		l_{\text{div}}(P_k) &= \sum_{c \in \{R,G,B\}} \sum_{i,j} \bigg( |\Delta_h P_{k,c}| + |\Delta_v P_{k,c}| + \\
		&\quad |\Delta_d P_{k,c}| + |\Delta_{ad} P_{k,c}| \bigg)
	\end{aligned}
	\]
	where $\Delta_h$, $\Delta_v$, $\Delta_d$, and $\Delta_{ad}$ denote first-order difference operators in the horizontal, vertical, diagonal, and anti-diagonal directions, respectively. Based on the descending order of $l_{\text{div}}$ values, we introduce a selection ratio $J\%$ to construct two views with distinctly different physical texture distributions: 
	the Texture-Rich View ($I_{\text{rich}}$) is reconstructed from the top $J\%$ high-complexity patches, concentrating high-frequency edges and grid artifacts; 
	the Texture-Poor View ($I_{\text{poor}}$) is reconstructed from the bottom $J\%$ low-complexity patches, focusing on low-frequency anomalies in smooth backgrounds. This process achieves physical decoupling of semantic content from texture statistical characteristics.
	
	\subsubsection{Residual Mapping and Compression}
	To expose faint underlying manipulation traces, the reconstructed views are fed into a Spatial Rich Model (SRM) layer \cite{Fridrich2012RichMF}. This layer applies 30 fixed high-pass filters to the RGB channels, directly mapping the input to a 90-channel high-dimensional residual feature space. Subsequently, a lightweight convolutional encoder (Conv-BN-ReLU) projects the sparse residuals into a compact 32-dimensional feature space. Ultimately, this module outputs semantic-agnostic texture features $F_{\text{rich}}, F_{\text{poor}} \in \mathbb{R}^{B \times 32 \times H \times W}$, which serve as the foundational input for the subsequent frequency-domain attention module. For the SRM layer, the 7 basic convolution kernels and their variants are presented in Appendix \ref{tab:conv_kernels}.
	
	\subsection{FFT-based Learnable Frequency Attention Module}
	Although spatial feature extraction can effectively capture local texture mutations, the artifacts introduced by generative models (especially transposed convolution-based GANs and denoising-based Diffusions) often manifest as global spectral statistical anomalies. These anomalies are easily obscured by complex semantic content in the spatial domain, but appear as prominent energy spikes or spectrum gaps in the frequency domain.
	
	To compensate for the inherent deficiency of pure spatial convolution in capturing long-range frequency dependencies, we propose a plug-and-play Learnable Frequency Attention (LFA) module. Unlike methods like FCANet that rely on manual frequency region division and fixed weighting, the core innovation of LFA lies in designing a "physical initialization-adaptive fine-tuning" masking mechanism. This mechanism leverages the global receptive field of FFT to decouple and enhance high-frequency artifacts in texture-rich regions and low-frequency anomalies in texture-poor regions, thereby realizing the recalibration of spatial features from the perspective of spectral energy perception.
	
	\subsubsection{Spectral Transformation \& Geometrically Constrained Masking}
	First, to project the feature representation from the spatial domain to the frequency domain, we perform 2D Fast Fourier Transform (2D FFT) on the input rich-texture feature $F_{\text{rich}}$ and poor-texture feature $F_{\text{poor}}$, respectively. Then, we apply the \text{FFTShift} operation to move the DC component to the center of the spectrum, obtaining the centered complex spectrograms $\mathcal{F}_{\text{rich}}$ and $\mathcal{F}_{\text{poor}}$:
	\[
	\mathcal{F}(u,v) = \text{FFTShift}\left(\text{FFT2}\left(F(x,y)\right)\right)
	\]
	
	FCANet \cite{Qin2020FcaNetFC} usually divides the spectrum into several fixed parts (e.g., only selecting the lowest-frequency DCT components), which struggles to adapt to the significant artifact distribution differences between different generative architectures (e.g., StyleGAN\cite{}\} vs. Stable Diffusion). To this end, we propose a two-branch learnable masking mechanism initialized based on Euclidean distance, embedding physical priors into the deep network:
	
	We first define a normalized Euclidean distance matrix $D$ from any coordinate $(u,v)$ in the spectrum to the center:
	\[
	D(u,v) = \frac{\sqrt{(u - H/2)^2 + (v - W/2)^2}}{R_{\text{max}}}, \quad D \in [0,1]
	\]
	where $R_{\text{max}}$ is the maximum Euclidean distance from the center to the corner. Based on this geometric constraint, we design physically inclined initialization masks for the two branches:
	\begin{enumerate}
		\item \textbf{Rich-texture High-frequency Preference Mask ($M_{\text{high}}^{\text{init}}$)}: \\
		Since generative models often introduce periodic gridding artifacts in complex texture regions due to upsampling, these artifacts are mainly concentrated in the high-frequency band. Therefore, we initialize the mask as 
		\[
		M_{\text{high}}^{\text{init}} = \alpha \cdot D(u,v) + \alpha
		\]
		so that its weight increases with the center distance, guiding the model to prioritize edge high-frequency information.
		
		\item \textbf{Poor-texture Low-frequency Preference Mask ($M_{\text{low}}^{\text{init}}$)}: \\
		Conversely, in smooth regions, generative models often fail to maintain the consistency of low-frequency energy (e.g., color block residue). Therefore, we initialize the mask as 
		\[
		M_{\text{low}}^{\text{init}} = \alpha \cdot (1 - D(u,v)) + \alpha
		\]
		forcing the model to focus on low-frequency anomalies in the center of the spectrum.
	\end{enumerate}
	
	\begin{figure}[!htbp] 
		\centering
		\includegraphics[width=\linewidth]{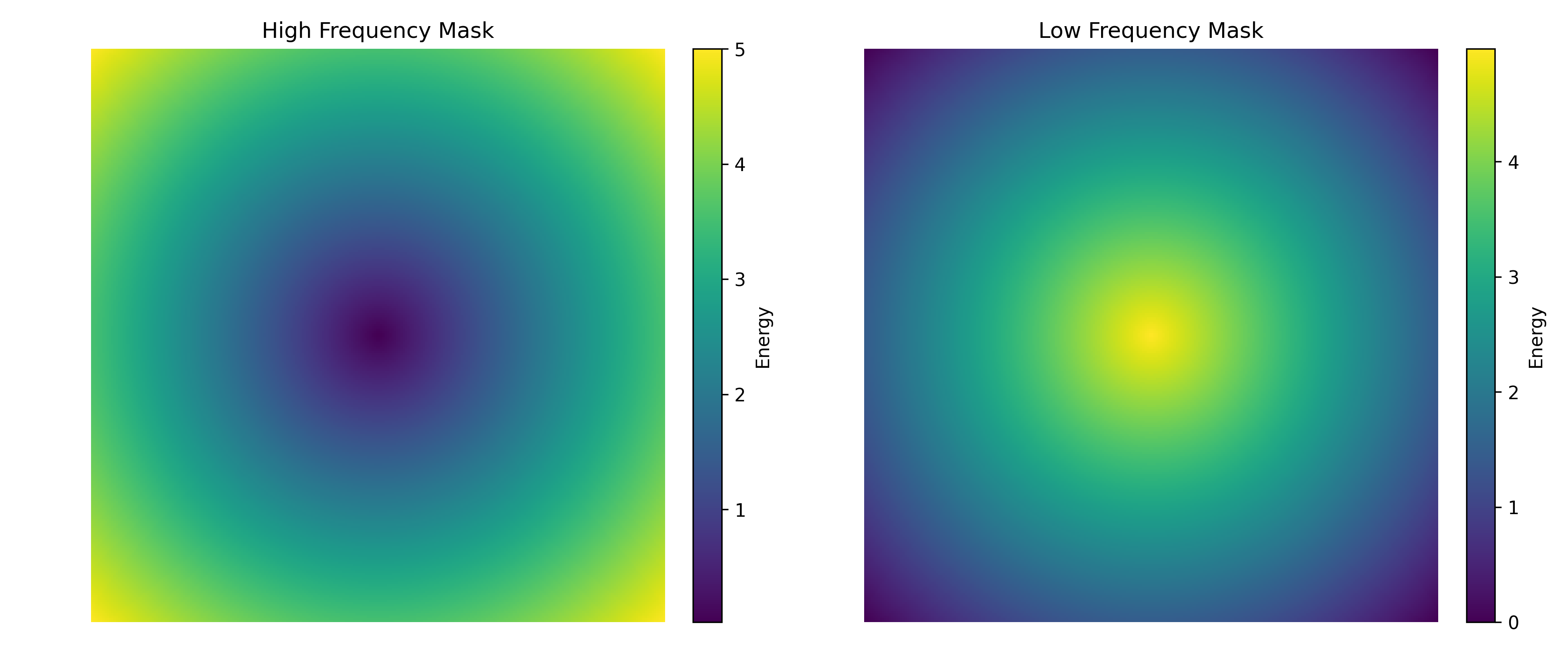}  
		
		\caption{Initialization Heatmap for High and Low Frequency Masks}
		\label{fig5}
	\end{figure}
	
	It should be emphasized that these masks $M_{\text{high}}$ and $M_{\text{low}}$ are fully learnable parameters. During training, the network can adaptively adjust the mask distribution according to the loss function, breaking through the physical constraints of initialization and capturing complex counterfeit fingerprints distributed across frequency bands. The application of the mask implements soft weighting via the Sigmoid activation function:
	\[
	\tilde{\mathcal{F}}_{\text{rich}} = \mathcal{F}_{\text{rich}} \odot \sigma(M_{\text{high}}), \quad \tilde{\mathcal{F}}_{\text{poor}} = \mathcal{F}_{\text{poor}} \odot \sigma(M_{\text{low}})
	\]

	\subsubsection{Group-wise Energy Aggregation \& Spatial Recalibration}
	To convert high-dimensional spectral information into effective guidance signals for spatial features and reduce computational redundancy between channels, we design a \textbf{group-wise energy aggregation} strategy.
	
	We divide the feature channel $C$ into $G$ subgroups (Groups), assuming that channels within the same group share similar frequency response patterns. For the $g$-th subgroup, we calculate the global spectral energy magnitude after mask screening:
	\[
	E_g = \sum_{u=0}^{H-1} \sum_{v=0}^{W-1} \left| \tilde{\mathcal{F}}_g(u,v) \right|
	\]
	This step is essentially a global pooling of "information density" within a specific frequency band. Then, to establish the mapping relationship between frequency and space, we introduce a learnable frequency-sensitive weight matrix $W_{freq} \in \mathbb{R}^G$. This weight matrix maps the scalar energy to channel attention coefficients and generates non-negative gain factors via ReLU activation.
	
	Finally, we project the global weights derived from the frequency domain back to the spatial domain, performing channel-wise adaptive recalibration on the original spatial features:
	\[
	F_{\text{rich}}' = F_{\text{rich}} \otimes \text{ReLU}(W_{\text{high}} \cdot E_{\text{rich}})
	\]
	\[
	F_{\text{poor}}' = F_{\text{poor}} \otimes \text{ReLU}(W_{\text{low}} \cdot E_{\text{poor}})
	\]
	where $\otimes$ denotes broadcast multiplication in the channel dimension. Through this mechanism, the LFA module not only utilizes the global receptive field of FFT to capture long-range cross-pixel dependencies, but also dynamically \textbf{amplifies} frequency bands containing prominent counterfeit traces while \textbf{suppressing} background frequency interference related to semantic content via the learnable masking strategy, thus enabling the detector to robustly extract universal counterfeit fingerprints in cross-model scenarios.
	
	\subsection{Cascaded Discriminative Module}
	The texture-rich features $F'_{\text{rich}}$ and texture-poor features $F'_{\text{poor}}$, enhanced by the LFA module, are concatenated along the channel dimension to form the fused feature map $F_{\text{fused}} \in \mathbb{R}^{B \times 64 \times H \times W}$. 
	
	To extract robust decision boundaries from the fused features, we design a Cascaded Discriminative Module (CDM). This module adopts a hierarchical Convolutional Neural Network architecture, composed of alternating convolutional blocks (Conv-BatchNorm-ReLU) and downsampling layers. As the network deepens and spatial dimensions reduce, the model progressively abstracts global forgery fingerprints from local texture anomalies. Finally, a Global Adaptive Average Pooling layer and a fully connected layer are used to output the final prediction score.
	
	The detailed structure is presented in Appendix \ref{tab:discriminator_design}: Design of the Cascaded Discrimination Module.
	
	\section{experimental results}
	\subsection{Experimental Setup}
	
	\subsubsection{Training Dataset}
	
\begin{figure*}[!htbp]
		\centering
		\includegraphics[width=17cm]{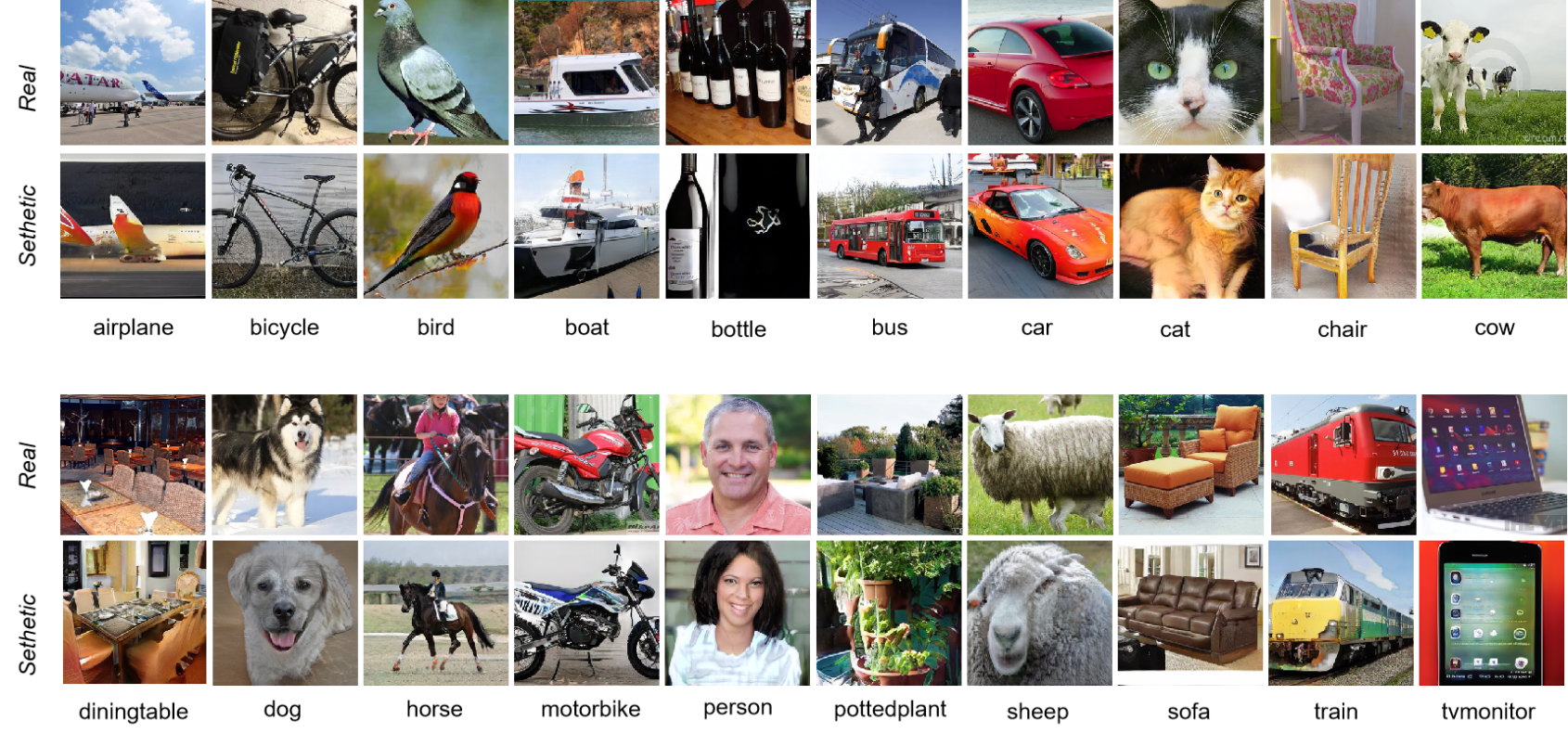}    
		\caption{The training set is sourced from ProGAN and comprises 20 categories. The above provides an overview of the image statistics for each category.}
		\label{dataset_constrution}
	\end{figure*}

	Following the generic paradigm of CNNSpot \cite{Wang_2020_CVPR}, we constructed a \textbf{single-source training} dataset to rigorously verify the cross-domain generalization capability of the model:
	\begin{itemize}
		\item \textbf{Data Composition \& Motivation}: We selected 360,000 real images covering diverse semantic scenes from the LSUN dataset \cite{2015arXiv150603365Y} as positive samples, and exclusively used 360,000 generated images from ProGAN \cite{2017arXiv171010196K} as negative samples. This training strategy, which solely leverages early-stage generative models, aims to compel the detector to capture intrinsic low-level defects of generative architectures rather than overfitting to dataset-specific semantic features. Effective detection of advanced models (e.g., StyleGAN or Diffusion) using only this training data serves as robust evidence of the model's strong generalization ability.(dataset constrution is shown in figure \ref{dataset_constrution})
		\item \textbf{Preprocessing}: A unified pipeline including random cropping, size normalization, and random JPEG compression was applied to all images to eliminate interference caused by variations in image size and compression quality.
	\end{itemize}
	
	\subsubsection{Cross-Model Benchmark}
	To comprehensively evaluate the model's robustness against unseen attacks, we adopted the comprehensive test benchmark AIGCDetectBenchmark \cite{2023arXiv231112397Z}, which encompasses 17 mainstream generative models categorized into four groups:
	\begin{itemize}
		\item \textbf{GANs Series}: Includes 7 classic models (ProGAN, StyleGAN, BigGAN, CycleGAN, StarGAN, GauGAN, StyleGAN2), used to evaluate the model's sensitivity to traditional generative artifacts.
		\item \textbf{Diffusion Models}: Includes 5 open-source diffusion models (Stable Diffusion v1.4/v1.5, SDXL, Glide, ADM, VQDM), representing a new generation of generative paradigms with fine-grained texture characteristics.
		\item \textbf{Domain-Specific Models}: Includes WFIR (Face Restoration), used to evaluate the model's effectiveness in specific facial forgery scenarios.
	\end{itemize}
	The 17 mainstream generative models are detailed in Table 1.

	\begin{table}[!htbp] 
		\centering 
		\caption{Detailed Information of Test Dataset} 
		\setlength{\tabcolsep}{3pt} 
		\scriptsize 
		\begin{tabular}{ccccc}
			\toprule
			Model Category & Gen Model & Size & Quantity & Source \\
			\midrule
			\multirow{7}{*}{\makecell[c]{GANs \\ Generator \\ (7)}} 
			& ProGAN & 256×256 & 8.0k & LSUN \\
			& StyleGAN & 256×256 & 12.0k & LSUN \\
			& BigGAN & 256×256 & 4.0k & ImageNet \\
			& CycleGAN & 256×256 & 2.6k & ImageNet \\
			& StarGAN & 256×256 & 4.0k & CelebA \\
			& GauGAN & 256×256 & 10.0k & COCO \\
			& Stylegan2 & 256×256 & 15.9k & LSUN \\
			\midrule
			Facial-Specific (1) & WFIR & 1024×1024 & 2.0k & FFHQ \\
			\midrule
			\multirow{4}{*}{\makecell[c]{Business \\ API \\ (4)}}
			& ADM & 256×256 & 12.0k & ImageNet \\
			& Glide & 256×256 & 12.0k & ImageNet \\
			& Midjourney & 1024×1024 & 12.0k & ImageNet \\
			& DALL-E2 & 256×256 & 2.0k & ImageNet \\
			\midrule
			\multirow{5}{*}{\makecell[c]{Diffusions \\ Generator \\ (5)}}
			& SDv1.4 & 512×512 & 12.0k & ImageNet \\
			& SDv1.5 & 512×512 & 16.0k & ImageNet \\
			& VQDM & 256×256 & 12.0k & ImageNet \\
			& Wukong & 512×512 & 12.0k & ImageNet \\
			& SDXL & 1024×1024 & 4.0k & COCO \\
			\bottomrule
		\end{tabular}
	\end{table}
	
	\subsection{Baseline Methods}
	To systematically evaluate the detection performance and generalization boundaries of the proposed method, this study selected 12 representative state-of-the-art detection methods as baselines, specifically including: CNNSpot (CVPR 2020) \cite{Wang_2020_CVPR}, Gram-Net (CVPR 2020) \cite{2020arXiv200200133L}, PatchCraft (arXiv 2023) \cite{2023arXiv231112397Z}, LNP (ECCV 2022) \cite{10.1007/978-3-031-19781-9_6}, LGrad (CVPR 2023) \cite{10203908}, FreDect (ICML 2020) \cite{10.5555/3524938.3525242}, FcaNet (ICCV 2021) \cite{Qin2020FcaNetFC}, Fusing (ICIP 2022) \cite{Ju2022FusingGA}, DIRE (ICCV 2023) \cite{2023arXiv230309295W}, FreqNet (AAAI 2024) \cite{2024arXiv240307240T}, UnivFD (CVPR 2023) \cite{2023arXiv230210174O}.
	
	\subsection{Implementation Details}
	We adopted the "Smash \& Reconstruction" strategy proposed by PatchCraft \cite{2023arXiv231112397Z}, where input images were randomly cropped into 192 patches of size $32 \times 32$. Based on four-directional pixel residual quantization, the top 33\% and bottom 33\% of patches with the highest and lowest texture complexity were selected, respectively, to reconstruct $256 \times 256$ texture-rich and texture-sparse images. 
	
	The Frequency Channel Attention (FCA) module was configured with 8 feature groups (4 channels per group) to achieve full spectral band coverage. The model was trained on NVIDIA 4090 GPUs using the PyTorch framework, with the Adam optimizer (initial learning rate: $1 \times 10^{-4}$, batch size: 32).
	\begin{table*}[!htbp] 
		\centering
		\caption{Experimental Software and Hardware Environment Configuration}
		\label{tab:env_config} 
		\setlength{\tabcolsep}{8pt} 
		\small 
		\begin{tabular}{cc}
			\toprule
			Configuration Item & Parameter Value \\
			\midrule
			Operating System & Ubuntu 20.04 \\
			CPU & Intel(R) Xeon(R) Gold 5418Y × 10 Cores \\
			GPU & RTX 4090 \\
			Available VRAM & 24564 MiB \\
			Programming Language & Python 3.8 \\
			Programming IDE & Visual Studio Code \\
			Deep Learning Framework & PyTorch 2.0.1 \\
			CUDA Version & CUDA 11.8 \\
			\bottomrule
		\end{tabular}
	\end{table*}
	To enhance robustness, online data augmentation was introduced with a trigger probability of 10\%, including random JPEG compression ($Q \in [70, 100]$) and Gaussian blur ($\sigma \in [0, 1]$).

	\subsection{Evaluation Metrics}
	Following the mainstream evaluation protocols in this field, this study adopts Accuracy (ACC) and Average Precision (AP) as core metrics to quantify the detection performance of the model:
	\begin{itemize}
		\item \textbf{Accuracy (ACC)}: Measures the overall prediction correctness of the model in the real/fake binary classification task, defined as the ratio of correctly classified samples to the total number of samples.
		\item \textbf{Average Precision (AP)}: Calculated as the area under the Precision-Recall (P-R) curve (with generated images treated as positive samples). Compared with ACC, AP can more comprehensively evaluate the model's performance across different confidence thresholds, and it exhibits stronger robustness and reference value especially for high-fidelity hard samples and class-imbalanced scenarios.
	\end{itemize}

	\subsection{Experimental Results}
	The evaluated average precision is presented in Table \ref{tab:accuracy_eval_final}, and the average precision evaluation results are shown in Table \ref{tab:ap_eval} (the best performance is \textbf{bolded}, and the second-best performance is underlined).
	
	\begin{table*}[h!t]
		\centering
		\caption{Accuracy Evaluation Results}
		\label{tab:accuracy_eval_final}
		\small
		\begin{tabularx}{\linewidth}{>{\centering\arraybackslash}m{1.2cm} >{\centering\arraybackslash}m{1cm} *{11}{>{\centering\arraybackslash}X}}
			\toprule
			Category & Generator & CNNS & FreDetc & Fusing & GramNet & LNP & LGard & D-G & D-D & UniFD & Pat.C. & OursFFT \\
			\midrule
			\multirow{7}{=}{\centering GANs (7)} 
			& ProGAN & \textbf{100.00} & 99.36 & \textbf{100.00} & 99.99 & 99.95 & 99.83 & 95.19 & 52.75 & 99.81 & \textbf{100.00} & 99.94 \\
			& StyleGAN & 90.17 & 78.02 & 85.20 & 87.05 & \underline{92.64} & 91.08 & 83.03 & 51.31 & 84.93 & \textbf{92.77} & 88.78 \\
			& BigGAN & 71.17 & 81.97 & 77.40 & 67.33 & 88.43 & 85.62 & 70.12 & 49.70 & \underline{95.08} & \textbf{95.80} & 87.58 \\
			& CGAN & 87.62 & 78.77 & 87.00 & 86.07 & 79.07 & 86.94 & 74.19 & 49.58 & \textbf{98.33} & 70.17 & \underline{91.22} \\
			& StarGAN & 94.60 & 94.62 & 97.00 & 95.05 & \textbf{100.00} & 99.27 & 95.47 & 46.72 & 95.75 & \underline{99.97} & 99.30 \\
			& GauGAN & 81.42 & 80.57 & 77.00 & 69.35 & 79.17 & 78.46 & 67.79 & 51.23 & \textbf{99.47} & 71.58 & \underline{78.83} \\
			& Stylegan2 & 86.91 & 66.19 & 83.30 & 87.28 & \textbf{93.82} & 85.32 & 75.31 & 51.72 & 74.96 & \underline{89.55} & 89.54 \\
			\midrule
			\centering Facial-Specific (1) 
			& WFIR & \textbf{91.65} & 50.75 & 66.80 & \underline{86.80} & 50.00 & 55.70 & 58.05 & 53.30 & 86.90 & 85.80 & 87.20 \\
			\midrule
			\multirow{4}{=}{\centering Business API (5)} 
			& ADM & 60.39 & 63.42 & 49.00 & 58.61 & \underline{83.91} & 67.15 & 75.78 & \textbf{98.25} & 66.87 & 82.17 & 85.58 \\
			& Glide & 58.07 & 54.13 & 57.20 & 54.50 & 83.50 & 66.11 & 71.75 & \textbf{92.42} & 62.46 & 83.79 & \underline{90.89} \\
			& Mjourney & 51.39 & 45.87 & 52.20 & 50.02 & 69.55 & 65.35 & 58.01 & \textbf{89.45} & 56.13 & \underline{90.12} & 88.68 \\
			& DALLE2 & 50.45 & 34.70 & 52.80 & 49.25 & 92.45 & 65.45 & 66.48 & 92.45 & 50.75 & \textbf{96.60} & \underline{96.35} \\
			\midrule
			\multirow{5}{=}{\centering Diffusions (5)} 
			& SDv1.4 & 50.57 & 38.79 & 51.00 & 51.70 & 89.33 & 63.02 & 49.74 & \underline{91.24} & 63.66 & \textbf{95.38} & 91.82 \\
			& SDv1.5 & 50.53 & 39.21 & 51.40 & 52.16 & 88.81 & 63.67 & 49.83 & \underline{91.63} & 63.49 & \textbf{95.30} & 91.39 \\
			& VODDM & 56.46 & 77.80 & 55.10 & 52.86 & 85.03 & 72.99 & 53.68 & \textbf{91.90} & \underline{85.31} & 88.91 & 85.21 \\
			& Wukong & 51.03 & 40.30 & 51.70 & 50.76 & 86.39 & 59.55 & 54.46 & \underline{90.90} & 70.93 & \textbf{91.07} & 88.76 \\
			& SDXL & 53.03 & 51.23 & 55.60 & 64.53 & 87.75 & 71.30 & 55.35 & \underline{91.28} & 50.73 & \textbf{98.43} & 97.18 \\
			\midrule
			\multicolumn{2}{c}{Average (Total 17)} 
			& 69.73 & 63.28 & 67.63 & 68.43 & 85.28 & 75.11 & 67.90 & 72.70 & 76.80 & \underline{89.85} & \textbf{90.49} \\
			\bottomrule
		\end{tabularx}
	\end{table*}
	

	\begin{table*}[h!t]
		\centering
		\caption{Average Precision Evaluation Results}
		\label{tab:ap_eval}
		\small 
		\begin{tabularx}{\linewidth}{>{\centering\arraybackslash}m{1.2cm} >{\centering\arraybackslash}m{1cm} *{11}{>{\centering\arraybackslash}X}}
			\toprule
			Category & Generator & CNNS & FreDect & Fusing & GramNet & LNP & LGard & D-G & D-D & UniFD & Pat.C. & OursFFT \\
			\midrule
			\multirow{7}{=}{\centering GANs (7)}
			& ProGAN & \textbf{100.00} & 99.99 & \textbf{100.00} & \textbf{100.00} & \textbf{100.00} & \textbf{100.00} & 99.08 & 58.79 & \textbf{100.00} & \textbf{100.00} & \textbf{100.00} \\
			& StyleGAN & \textbf{99.83} & 88.98 & \underline{99.50} & 99.23 & 99.27 & 98.31 & 91.74 & 56.68 & 97.56 & 98.96 & 97.86 \\
			& BigGAN & 85.99 & 93.62 & 90.70 & 81.79 & 94.54 & 92.93 & 75.25 & 46.91 & \underline{99.27} & \textbf{99.42} & 95.07 \\
			& CGAN & 94.94 & 84.78 & 95.50 & 95.33 & 89.52 & 95.01 & 80.56 & 50.03 & \textbf{99.80} & 85.26 & \underline{96.37} \\
			& StarGAN & 99.04 & 99.49 & 99.80 & 99.23 & \textbf{100.00} & \textbf{100.00} & 99.34 & 40.64 & 99.37 & \textbf{100.00} & \underline{99.93} \\
			& GauGAN & 90.82 & 82.84 & 88.30 & 84.99 & 84.54 & 95.43 & 72.15 & 47.34 & \textbf{99.98} & 81.33 & \underline{84.98} \\
			& Stylegan2 & 99.48 & 82.54 & \underline{99.60} & 99.11 & \textbf{99.70} & 97.89 & 88.29 & 58.03 & 97.90 & 97.74 & 97.80 \\
			\midrule
			\centering Facial-Specific (1)
			& WFIR & \textbf{99.85} & 55.85 & 93.30 & 95.21 & 42.75 & 57.99 & 60.13 & 59.02 & \underline{96.73} & 95.26 & 95.87 \\
			\midrule
			\multirow{4}{=}{\centering Business API (5)}
			& ADM & 75.67 & 61.77 & 94.10 & 73.11 & 93.37 & 72.95 & 85.84 & \textbf{99.79} & 86.81 & 93.40 & \underline{94.88} \\
			& Glide & 72.28 & 52.92 & 77.50 & 66.76 & 92.76 & 80.42 & 78.35 & \textbf{99.54} & 83.81 & 94.04 & \underline{97.19} \\
			& Mjourney & 66.24 & 46.09 & 70.00 & 56.82 & 86.92 & 71.86 & 61.86 & \textbf{97.32} & 74.00 & \underline{96.48} & 95.77 \\
			& DALLE2 & 53.51 & 38.20 & 68.12 & 49.82 & 98.26 & 82.55 & 74.48 & \textbf{99.71} & 63.04 & \underline{99.56} & 99.00 \\
			\midrule
			\multirow{5}{=}{\centering Diffusions (5)}
			& SDv1.4 & 61.20 & 37.83 & 65.40 & 59.83 & 96.34 & 62.37 & 49.87 & \underline{98.61} & 86.14 & \textbf{99.06} & 97.26 \\
			& SDv1.5 & 61.56 & 37.76 & 65.70 & 60.37 & 96.00 & 62.85 & 49.52 & \underline{98.83} & 85.64 & \textbf{99.06} & 97.31 \\
			& VQDM & 68.83 & 85.10 & 75.60 & 61.13 & 94.91 & 77.47 & 54.57 & \textbf{98.98} & \underline{96.53} & 96.26 & 92.81 \\
			& Wukong & 57.34 & 39.58 & 64.60 & 55.62 & 95.33 & 62.48 & 55.38 & \underline{98.37} & 91.07 & \textbf{97.54} & 94.91 \\
			& SDXL & 72.62 & 49.45 & 79.41 & 68.22 & 87.75 & 80.03 & 53.97 & \underline{99.10} & 67.59 & \textbf{99.89} & 99.61 \\
			\midrule
			\multicolumn{2}{c}{Average (Total 17)}
			& 79.95 & 66.87 & 83.95 & 76.86 & 91.29 & 81.80 & 72.38 & 76.92 & 89.73 & \underline{96.07} & \textbf{96.27} \\
			\bottomrule
		\end{tabularx}
	\end{table*}

	\begin{table}[htbp]
		\centering
		\caption{Model Parameter Comparison}
		\label{tab:model_params}
		\begin{tabular*}{\linewidth}{@{\extracolsep{\fill}}ccc@{}}
			\toprule
			Model     & Params (M) & Weight Size (MB) \\ 
			\midrule
			VGG-16    & 138.36     & 527.80           \\
			ResNet-18 & 11.69      & 44.66            \\
			ResNet-34 & 21.80      & 83.27            \\
			ResNet-50 & 25.56      & 97.79            \\
			Ours-FFT  & \underline{1.20}  & \underline{13.85}\\
			\bottomrule
		\end{tabular*}
	\end{table}

	\subsection{Experimental Results Analysis}
	From the experimental results, the proposed method demonstrates excellent detection performance across multiple scenarios: On test sets of generation models such as CycleGAN, StarGAN, and StyleGAN2, its detection accuracy ranks \textit{second}, showing good adaptability to mainstream generation models; In highly challenging test scenarios like Midjourney and DALL-E 2, the accuracy of the proposed method remains \textit{first}. Most critically, the proposed method achieves an average accuracy of 90.49\%, which is higher than the average accuracy of all baselines. This result fully verifies the strong generalization ability of the proposed method for unknown generation models, and confirms the effectiveness of the core design of our method in capturing cross-model universal artifact features.
	
	Meanwhile, the proposed detection method shows good detection accuracy in multi-generation model test scenarios: On test sets of mainstream generation models (representing GANs and diffusion models, such as BigGAN, CycleGAN, and ADM), the detection accuracy of the proposed method ranks \textit{second}, which can reflect the wide adaptability of the method to different types of generated images; In terms of the average accuracy index, the proposed method ranks \textit{first} with a value of 96.27\%. This result fully verifies the rationality and effectiveness of the core design of the method, which can provide reliable technical support for subsequent cross-generation model image detection tasks.

	\subsection{Robustness Experiments}
	Various degradation distortions in real network propagation will destroy the subtle artifacts of generated images and greatly increase the difficulty of AI-generated image detection. Therefore, robustness is a core indicator to verify the practical application value of the model, and it is also the key to distinguishing performance between laboratory and real scenarios.
	
	This experiment refers to the general paradigm of related studies [11, 25, 47] to set degradation parameters, so as to quantify the robustness of the proposed method and baseline methods, and ensure the reproducibility and comparability of the experiment. The configurations of each degradation scenario are as follows:
	\begin{itemize}
		\item JPEG compression: Adopt the mainstream standard, set the quality factor (QF) to 95, to simulate the conventional compression processing of social platforms and storage tools;
		\item Gaussian blur: Use a Gaussian filter kernel for blur processing, set the kernel standard deviation ($\sigma$) to 1, to simulate detail blur caused by noise interference or low-resolution display during image transmission;
		\item Downsampling: Use bilinear interpolation to scale the image, set the scaling factor ($r$) to 0.5, to simulate the reduction of image resolution in scenarios such as mobile terminal adaptation and low-bandwidth transmission.
	\end{itemize}
	
	The average accuracy of each model in robustness detection is shown in Table \ref{tab:robustness_accuracy} (the best result in each method is \textbf{bolded}, and the second-best result is underlined).
	
	\begin{table*}[h!t] 
		\centering
		\caption{Average Accuracy of Each Model in Robustness Detection}
		\label{tab:robustness_accuracy} 
		\small 
		\setlength{\tabcolsep}{12pt} 
		\begin{tabular}{lccc} 
			\toprule
			Detection Method & JPEG Compression (QF=95) & Gaussian Blur ($\sigma$=1) & Downsampling ($r$=0.5) \\
			\midrule
			CNNSpot & 63.45 & 67.13 & 66.84 \\
			FreDect & 66.70 & 65.42 & 35.45 \\
			Fusion & 61.76 & 67.54 & 66.62 \\
			GramNet & 52.29 & 65.97 & 65.78 \\
			LNP & 53.56 & 65.88 & 63.28 \\
			LGrad & 52.89 & 72.22 & 62.54 \\
			DIRE-G & 66.49 & 64.00 & 56.09 \\
			DIRE-D & 70.27 & 70.46 & 62.26 \\
			UnivFD & 70.26 & 69.19 & 68.71 \\
			Patchcraft$^{[10]}$ & \underline{72.01} & \textbf{78.34} & \underline{75.92} \\
			\textbf{Ours-FFT} & \textbf{73.31} & \underline{85.03} & \textbf{76.13} \\
			\bottomrule
		\end{tabular}
	\end{table*}

	Based on the quantitative analysis of the experimental results, the proposed AIGC image detection method demonstrates excellent anti-distortion detection performance under multiple distortion interference scenarios:
	\begin{itemize}
		\item JPEG compression scenario: The average detection accuracy of the proposed method ranks \textit{first} among all baseline methods, which is 1.30\% higher than that of Patchcraft (ranked second); Compared with the distortion-free test scenario, the accuracy of the method in this scenario decreases by 17.18\%, but it still maintains the current optimal anti-compression distortion capability.
		\item Gaussian blur scenario: The average detection accuracy of the proposed method ranks \textit{second}, which is only 0.52\% lower than that of the DCT variant of the proposed method; Compared with the distortion-free test scenario, the accuracy only decreases by 4.94\%, and the performance attenuation amplitude is significantly lower than that in the JPEG compression scenario.
		\item Downsampling scenario: The proposed method ranks \textit{first} with an accuracy of 76.13\%, which is 0.21\% higher than that of PatchCraft (ranked second); Although the comprehensive performance of the method in this scenario is slightly inferior to PatchCraft, it is still significantly better than most baseline methods.
	\end{itemize}
	
	The above results can be attributed to the core design of the proposed method—the design of frequency-domain grouped attention weighted features for rich and sparse texture. This feature focuses on the essential characteristics of the frequency domain, and can screen key frequency bands through the grouped attention mechanism, effectively avoiding the interference of high-frequency compression noise; Even if the subtle artifacts in the spatial domain are destroyed by distortion operations such as JPEG compression, Gaussian blur, and downsampling, this feature can still capture the essential frequency-domain differences between real images and generated images.
	
	Overall, the proposed method maintains high detection accuracy and low performance attenuation in the three degradation scenarios, and can accurately capture the essential frequency-domain differences between real and generated images, confirming that the method has good anti-distortion detection capability.
	
	\subsection{Ablation Experiments}
	The Learnable Frequency Attention (LFA) module significantly enhances the model's cross-domain generalization capability through adaptive weighting of texture-rich and texture-sparse features. To validate its core value and determine the optimal configuration, this section focuses on analyzing the effectiveness of LFA components and the impact of group numbers on experimental results.
	
	\subsubsection{Differential Contributions of High- and Low-Frequency Attention Components}
	To clarify the specific roles of the high-frequency and low-frequency components within the LFA module, we designed three groups of comparative experiments and performed quantitative evaluations on a test set covering 17 mainstream generative models (detailed results are shown in Table 7):
	\begin{itemize}
		\item \textbf{w/o LFA}: Completely removing the frequency-domain attention module (baseline control).
		\item \textbf{w/o Low-Freq}: Retaining high-frequency weighting while removing only the low-frequency feature attention module.
		\item \textbf{w/o High-Freq}: Retaining low-frequency weighting while removing only the high-frequency feature attention module.
	\end{itemize}
	
	\begin{enumerate}
		\item \textbf{\textit{Low-Frequency Attention Module}}
		
		Experimental data shows that completely removing the LFA module leads to a significant 4.7\% drop in the model's average accuracy across the 17 test sets. Notably, when only the low-frequency weighting module is independently removed, the performance degradation further expands to 5.37\%, even exceeding the loss caused by removing the entire module.
		
		This counterintuitive phenomenon reveals the critical status of the low-frequency attention module: the frequency-domain anomalies in sparse texture regions captured by the low-frequency module are inherent, semantic-agnostic low-level defects of generative models. Since generative models struggle to perfectly replicate the hardware-dominated low-frequency statistical patterns of real images, these discrepancies constitute the most stable discriminative signals. The model's robustness relies on the complementary synergy between high-frequency and low-frequency features. Removing the low-frequency module not only results in the loss of global anomaly information but also disrupts the balance between features, leading to severe impairment of the model's generalization capability.
		
		\item \textbf{\textit{High-Frequency Attention Module}}
		
		In contrast, removing only the texture-rich high-frequency weighting module causes a mere 0.21\% slight drop in the model's average accuracy. This indicates that in most general scenarios, high-frequency information in texture-rich regions is not the core dependency for discrimination. The main reason is that high-frequency artifacts of generative models are easily masked by complex image textures, and their spectral distributions highly overlap with those of real images, resulting in low discriminability.

		However, in the specialized test on CycleGAN-generated images, removing the high-frequency module leads to a significant performance decline. This is because CycleGAN's local style transfer and detail reconstruction mechanisms introduce unique high-frequency synthetic traces in texture-rich regions. Therefore, although the high-frequency module makes a limited contribution in general scenarios, it provides irreplaceable supplementary value for such specific generative mechanisms.
	\end{enumerate}
	
	\subsubsection{Impact of Group Number on Feature Learning Mechanisms}
	To investigate the non-linear impact of feature group numbers on detection performance, we designed parameter sensitivity experiments and analyzed the underlying mechanisms by visualizing the difference heatmaps between the "learned mask" and the "initial mask" in the FFT weighting module (Figure 6). Experimental results (see Appendix 10) indicate:
	\begin{itemize}
		\item \textbf{8-Group Configuration}: When the number of groups is set to 8, the model achieves the peak average accuracy. The frequency-domain heatmap shows the broadest coverage of mask differences, indicating that the model can synergistically capture global high, medium, and low-frequency features as well as local details, achieving an optimal global-local balance in feature learning.
		\item \textbf{16-Group Configuration}: Increasing the number of groups to 16 causes the accuracy to drop to the lowest point. The heatmap reflects that the model excessively focuses on local frequency-domain features, leading to "missing the forest for the trees"—it struggles to extract complete global frequency patterns and falls into local overfitting.
		\item \textbf{32-Group Configuration}: Further increasing the number of groups to 32 results in constant differences in the high-frequency mask of texture-rich regions. This indicates that the model's learning of high-frequency features in these regions completely fails, forcing it to rely solely on information from texture-sparse regions, which severely limits the comprehensiveness of feature expression.
	\end{itemize}
	
\begin{figure}[!htbp]
		\centering
		\vspace{-5pt}
		\includegraphics[width=0.9\linewidth]{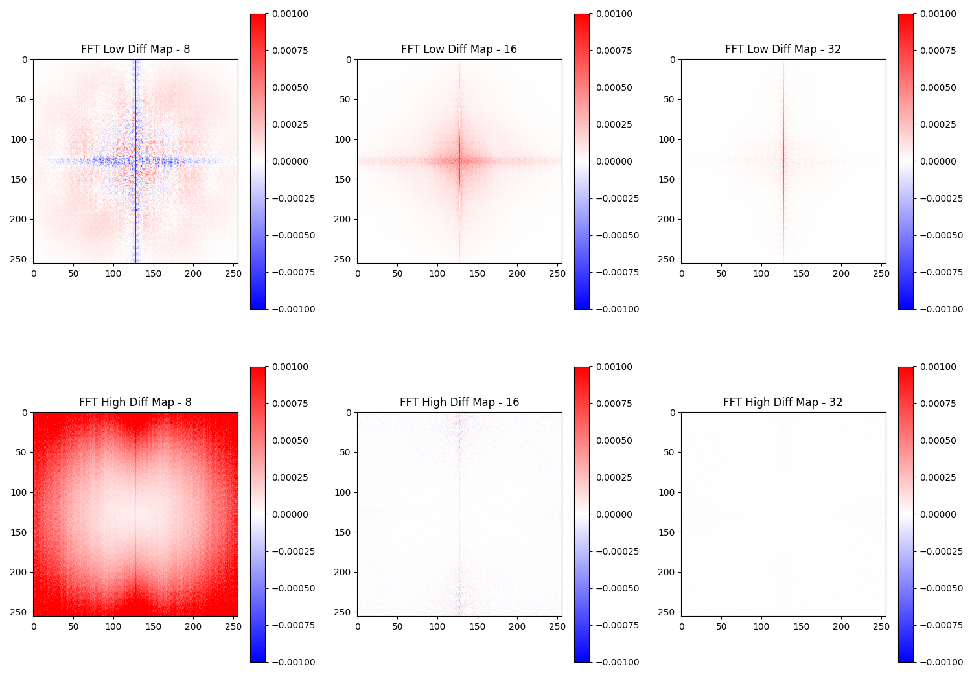}
		\vspace{-3pt}
		\caption{Difference heatmap between the learned mask and initial mask in the FFT weighting module under different group configurations.}
		\label{fig:group_mask_heatmap} 
	\end{figure}

	In summary, both quantitative data and visualization results (As shown in Figure \ref{fig:group_mask_heatmap}) collectively confirm that the 8-group configuration is the optimal choice for balancing global pattern capture and local detail learning. Excessive grouping fragments feature learning, disrupting this synergistic mechanism and ultimately leading to dual degradation of generalization performance and detection accuracy.
	
	\begin{table*}[h!t]
		\centering
		\caption{Evaluation of Ablation Experiments for LFA Module}
		\label{tab:ablation_lfa} 
		\setlength{\tabcolsep}{4pt}
		\begin{tabular*}{\textwidth}{@{\extracolsep{\fill}}clccc@{}}
			\toprule
			\makecell{Model \\ Category} & \makecell{Generative\\Model} & \makecell{w/o LFA \\ (Baseline)} & \makecell{w/o Low-Freq \\ Block} & \makecell{w/o High-Freq \\ Block} \\
			\midrule
			\multirow{7}{*}{\makecell{GANs \\ (7)}} 
			& ProGAN       & \textbf{99.96} & \underline{99.84} & 99.75 \\
			& StyleGAN     & 88.69 & \textbf{94.06} & \underline{91.21} \\
			& BigGAN       & \underline{78.50} & \textbf{82.63} & 78.33 \\
			& CycleGAN     & \underline{69.04} & \textbf{77.55} & 69.04 \\
			& StarGAN      & 95.20 & \textbf{99.65} & \underline{99.52} \\
			& GauGAN       & 72.09 & 65.24 & \textbf{86.84} \\
			& StyleGAN2    & 86.63 & \textbf{95.84} & \underline{92.00} \\
			\midrule
			\makecell{Facial- \\ Special \\ (1)} 
			& WFIR         & \underline{88.30} & 71.00 & \textbf{97.05} \\
			\midrule
			\multirow{4}{*}{\makecell{Business \\ API \\ (4)}} 
			& ADM          & 79.10 & \textbf{86.46} & \underline{84.80} \\
			& Glide        & 88.78 & 84.63 & \textbf{89.98} \\
			& Midjourney   & 70.84 & \underline{82.37} & \textbf{91.34} \\
			& DALLE2       & \underline{96.10} & 87.90 & \textbf{97.70} \\
			\midrule
			\multirow{5}{*}{\makecell{Diffusions \\ (5)}} 
			& SDv1.4       & 90.97 & 82.65 & \textbf{92.16} \\
			& SDv1.5       & \underline{91.16} & 82.44 & \textbf{91.87} \\
			& VODv1        & 81.31 & \underline{84.08} & \textbf{84.96} \\
			& Wukong       & 86.51 & 81.47 & \textbf{89.72} \\
			& SDXL         & \underline{95.30} & 89.35 & \textbf{98.40} \\
			\midrule
			\multirow{2}{*}{\makecell{Total \\ 17}} 
			& Avg. Acc. & 85.79 & \underline{85.12} & \textbf{90.28} \\
			& Decrease  & \underline{-4.70} & \textbf{-5.37} & -0.21 \\
			\bottomrule
		\end{tabular*}
	\end{table*}
	
	\subsection{T-SNE Analysis}
	To intuitively verify the model's ability to distinguish features of AIGC-generated images from different paradigms, this study designed a T-SNE dimensionality reduction visualization experiment \cite{2008Visualizing}. The last fully connected layer of the discriminator was replaced with an identity mapping layer to obtain the original high-dimensional discriminative feature vectors without classification mapping. T-SNE was then used to reduce the dimensionality of these vectors to a two-dimensional space, and the effectiveness of the model's feature extraction was demonstrated through the feature distribution in the low-dimensional space. A visualization dataset was constructed by selecting four types of samples covering two mainstream generation paradigms: GAN and Stable Diffusion.
	
\begin{figure*}[htbp]  
		\centering  
		\adjustbox{max width=\textwidth}{
			\begin{tabular}{ccc}  
				\subfigure[]{\includegraphics[width=0.32\textwidth]{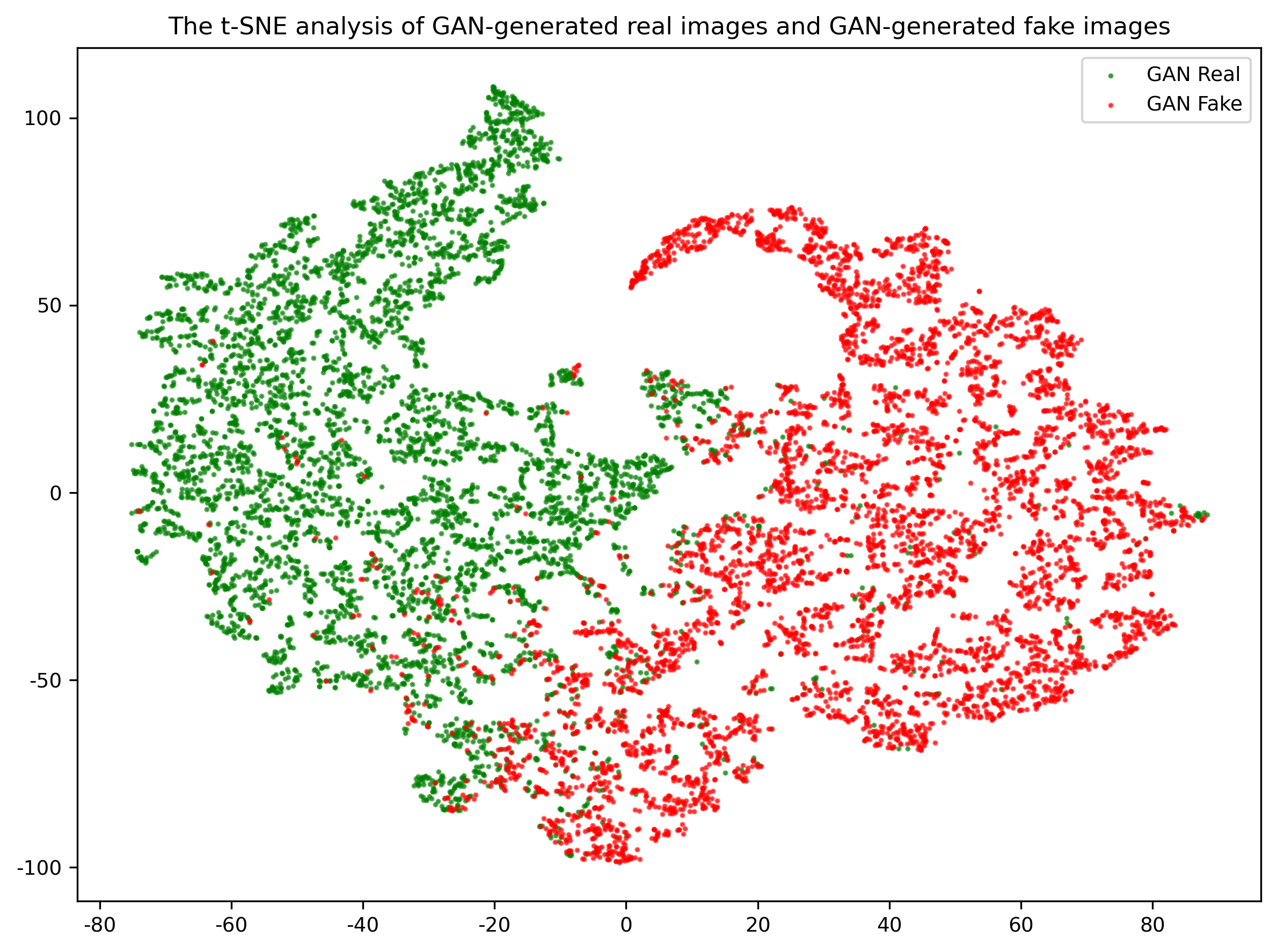}} &
				\subfigure[]{\includegraphics[width=0.32\textwidth]{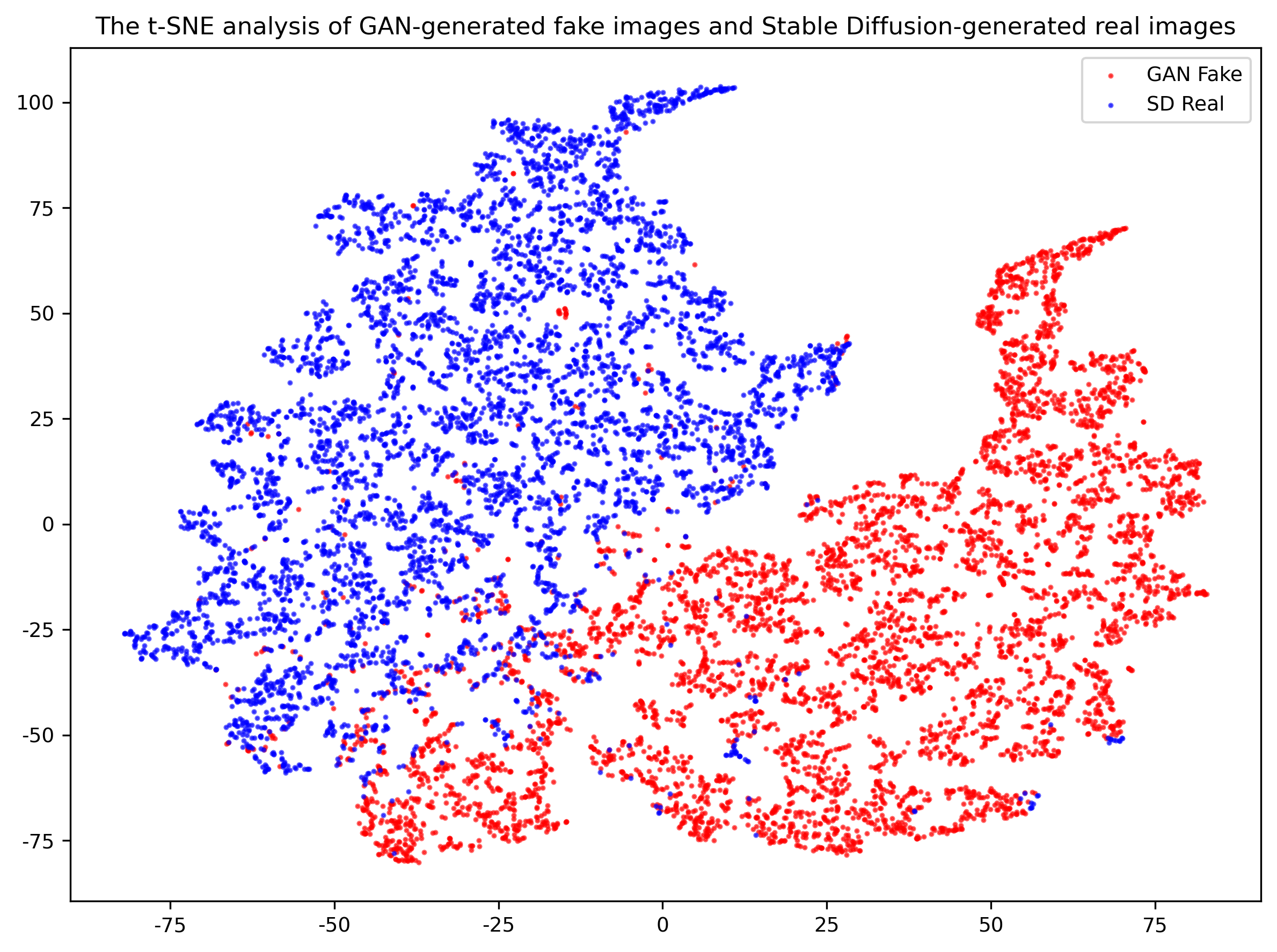}} &
				\subfigure[]{\includegraphics[width=0.32\textwidth]{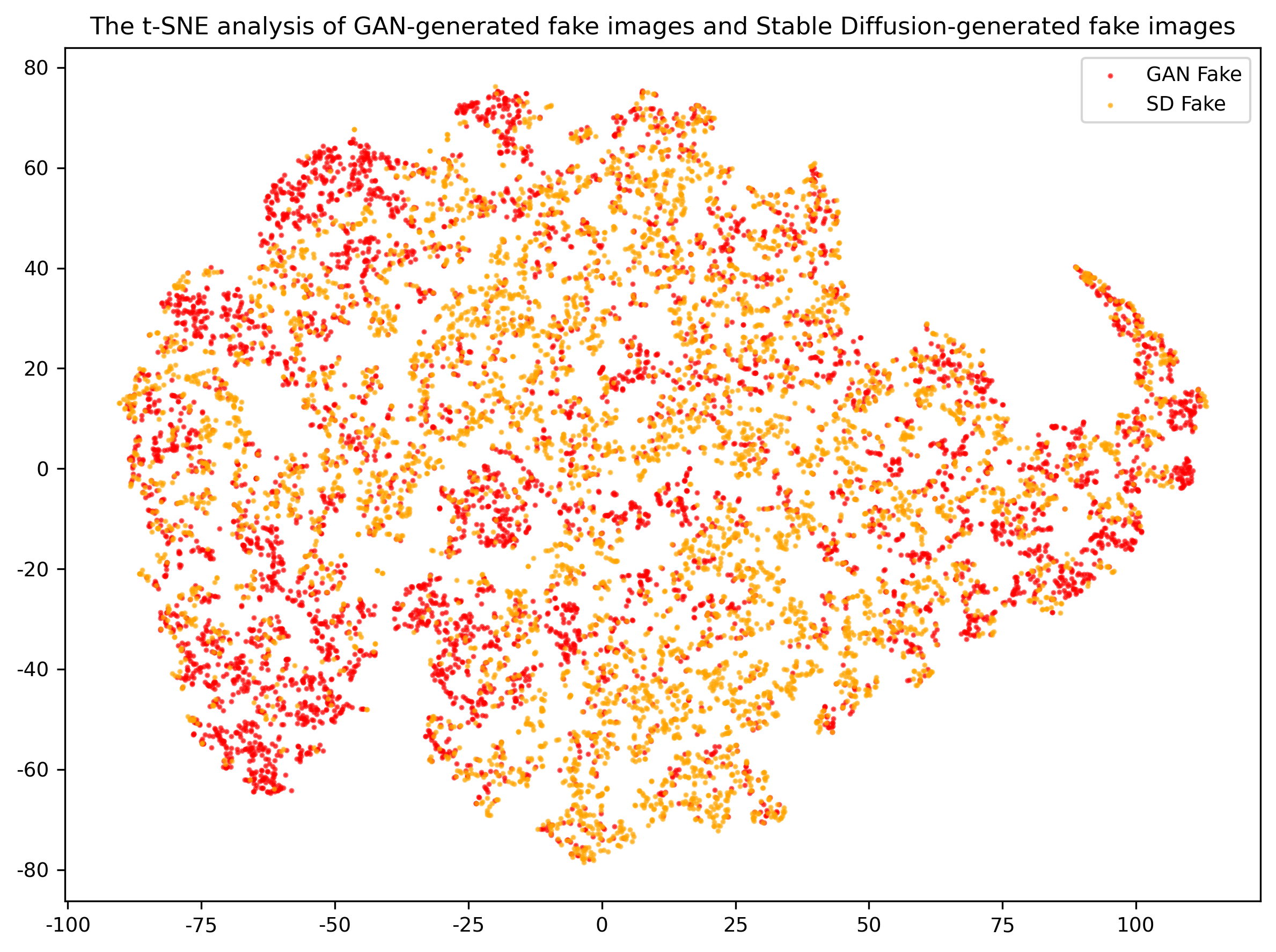}} \\
				\subfigure[]{\includegraphics[width=0.32\textwidth]{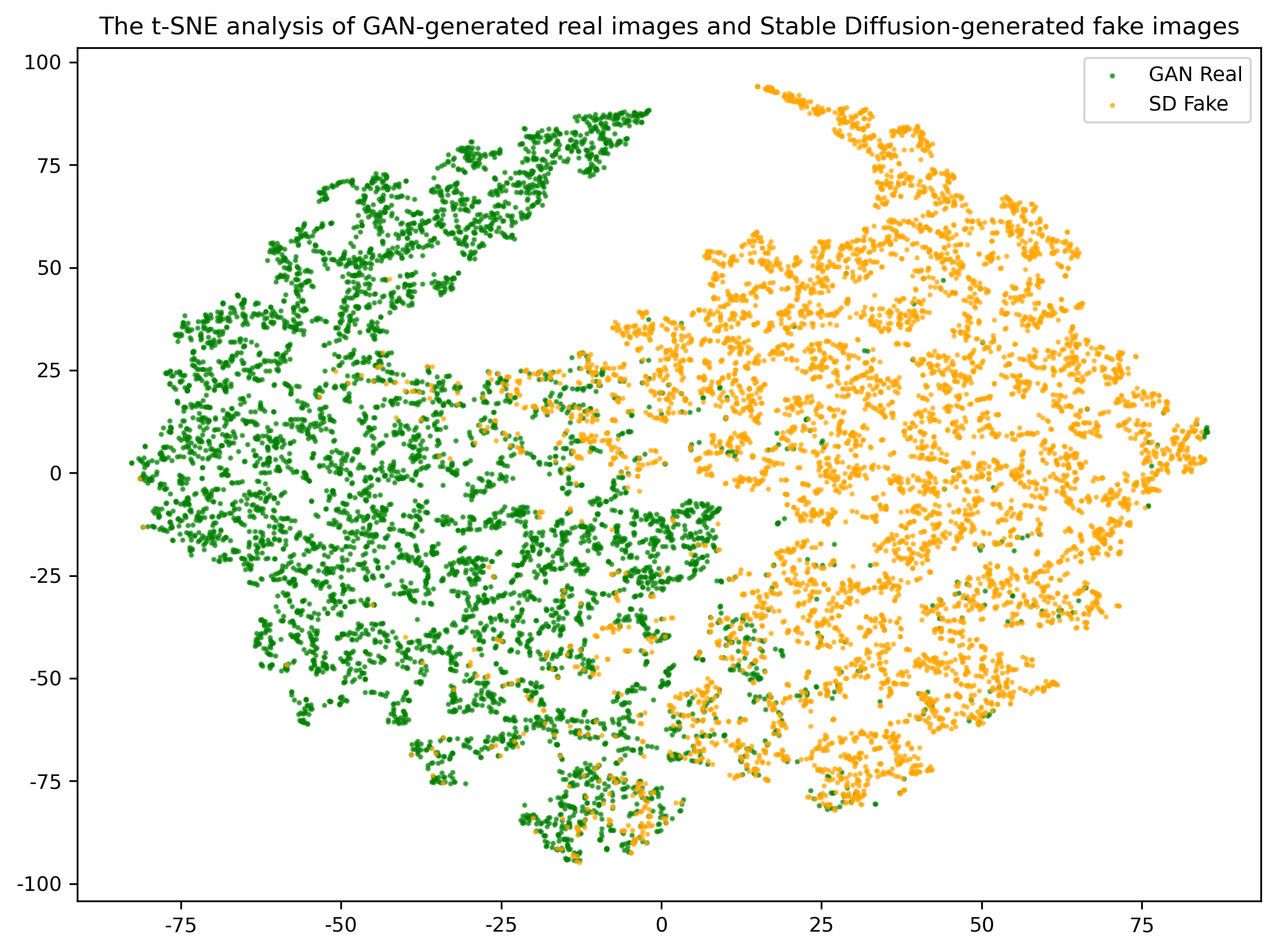}} &
				\subfigure[]{\includegraphics[width=0.32\textwidth]{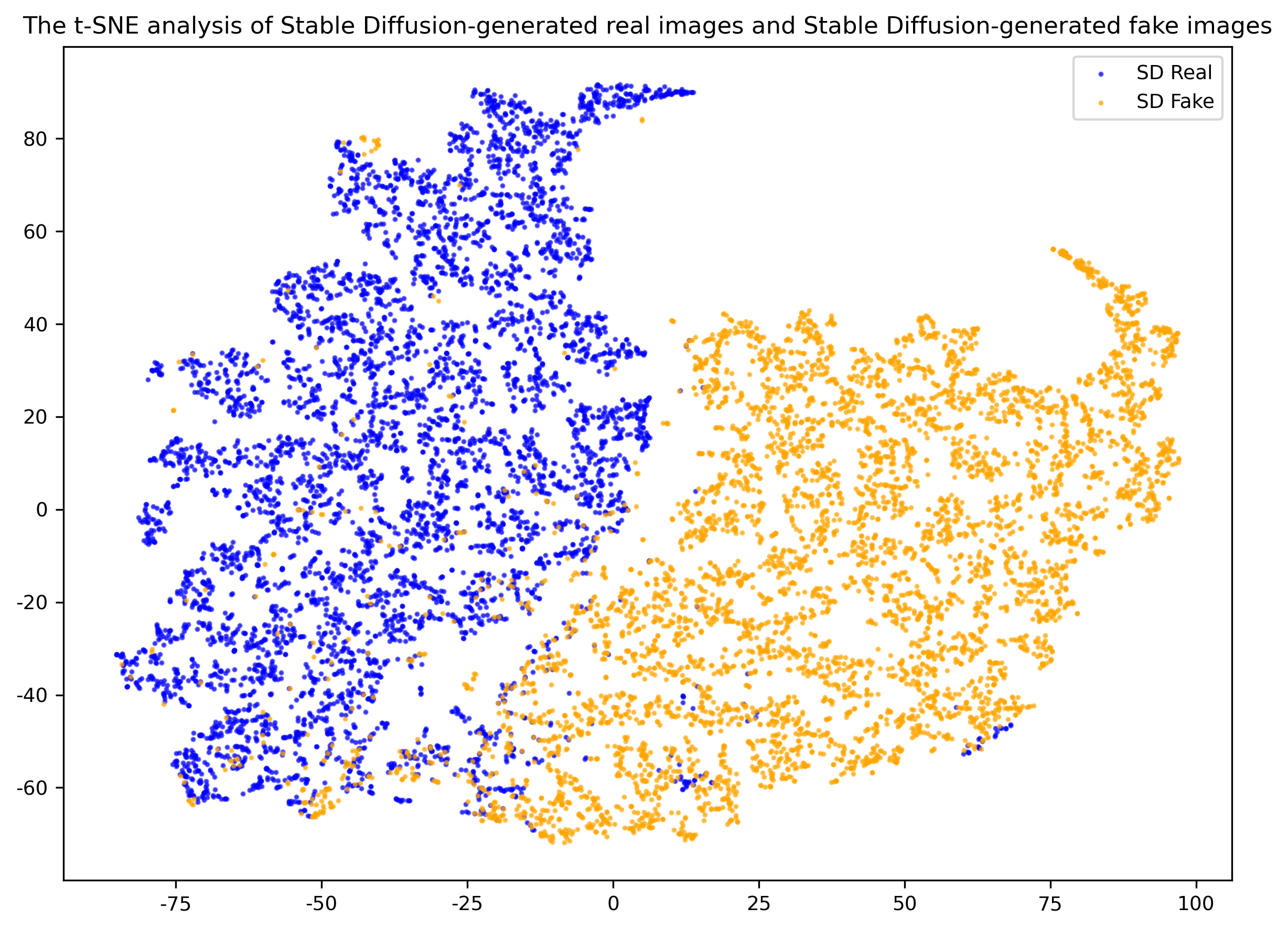}} &
				\subfigure[]{\includegraphics[width=0.32\textwidth]{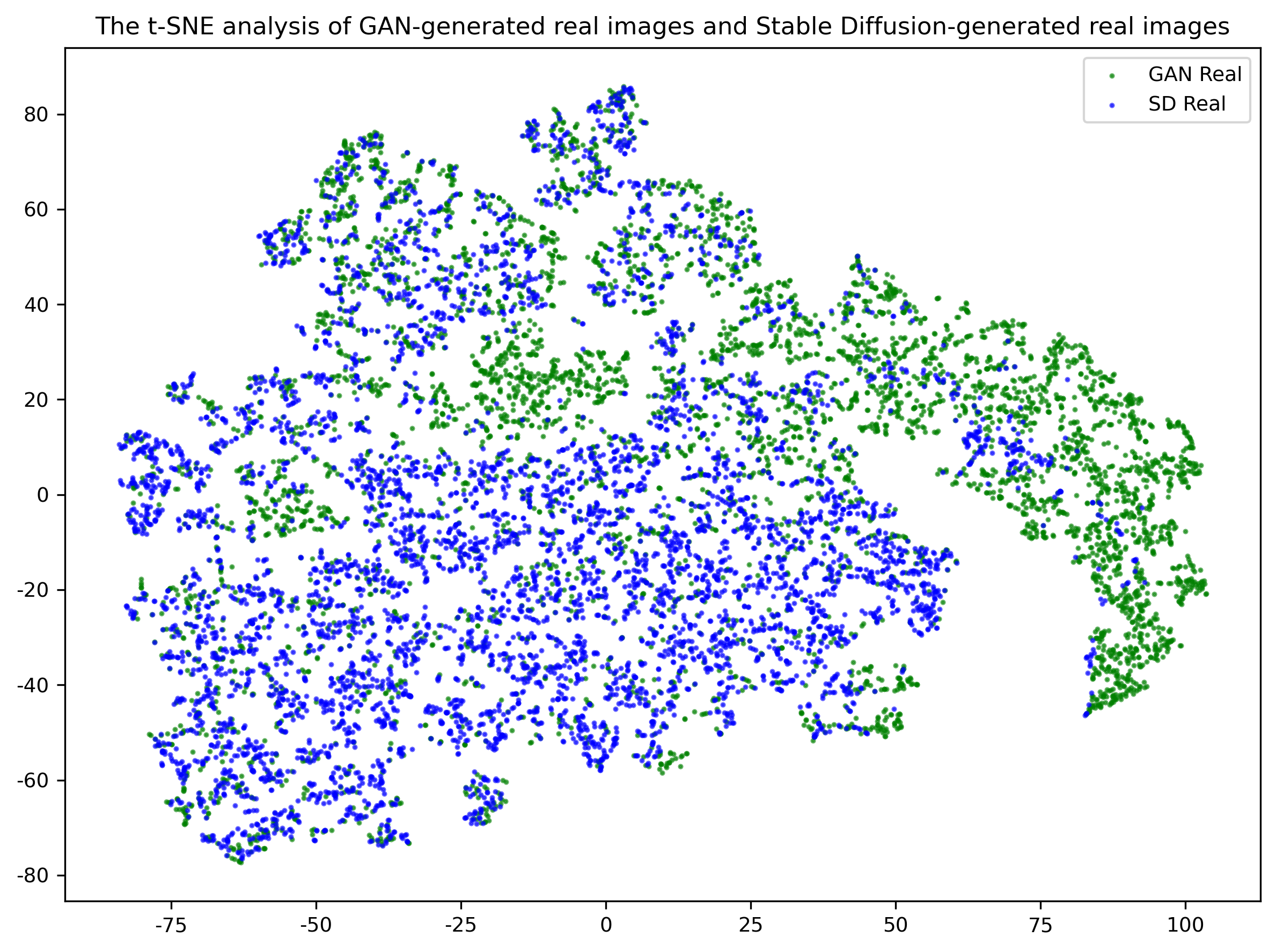}}
			\end{tabular}
		}
		\caption{t-SNE visualization of different feature distributions in AIGC detection}
		\label{fig:tsne_all}
	\end{figure*}

	The T-SNE visualization based on FFT weighting is shown in Figure \ref{fig:tsne_all}.

	The t-SNE visualization demonstrates superior clustering and distinct feature distributions. As shown in Figs. a, b, d, and e, the clear separation boundaries between real and fake images in both intra- and cross-architecture scenarios prove that the model captures universal underlying statistical anomalies rather than dataset-specific semantic biases, confirming its exceptional generalization across diverse AIGC models. Notably, the significant overlap and blurred boundaries between GAN and SD fake images in Fig. f reveal that the model has successfully identified deep-seated "universal fingerprints" that transcend specific synthesis paradigms. Furthermore, the highly intertwined and nearly identical distribution of real images from different sources in Fig. c validates the consistency of natural physical statistical laws, providing a stable "real-world prior" that ensures robust detection performance even against high-fidelity generative content.
	
	The visualization results indicate that the high-dimensional features of generated and real images exhibit excellent clustering effects after dimensionality reduction via t-SNE. As shown in \textbf{Figs. a, b, d, and e}, there are significant separation boundaries between real and fake images regardless of intra-architecture or cross-architecture comparisons. This high separability confirms that the model does not capture semantic biases specific to a particular dataset, but accurately extracts the universal underlying statistical artifacts commonly present in AIGC-generated images, verifying the model's outstanding generalization ability in complex cross-model detection tasks. Notably, in \textbf{Fig. f}, the feature points of fake images generated by two heterogeneous algorithms (GAN and SD) show extensive overlap with blurred boundaries, further revealing that the model has successfully mined deep-seated universal "fingerprints" that transcend specific generation paradigms, reflecting its acute capture of the intrinsic flaws in the generation mechanism. In addition, \textbf{Fig. c} shows that the features of real images from different sources are highly intertwined and overlapping in distribution, which experimentally confirms that natural images follow consistent physical statistical laws. This provides a reliable discriminative benchmark for the model, ensuring its robustness when facing high-fidelity generated content.
	
	The visualization results also show that in the boundary between real and generated images identified by the proposed model, a small amount of overlap exists between some generated images and real images. This phenomenon provides an experimental direction for subsequent model optimization.  
	
	\section{Conclusion}
	This paper focuses on solving the poor generalization problem of existing AI-generated image detection methods. The core finding is that cumulative upsampling operations in generative models lead to unique frequency-domain artifacts, which show distinct characteristics in texture-rich and texture-poor regions.
	
	Based on this insight, we clarify the essential feature differences between real and generated images and their formation mechanisms, innovatively design a learnable frequency attention mechanism for universal frequency-domain feature enhancement, and propose the universal detection method S$^2$F-Net. Experimental results verify the effectiveness of the method: trained only on ProGAN data, it achieves an average accuracy of 90.6\% across 17 generative models (overall accuracy 90.49\%), effectively overcoming the reliance of traditional methods on single-model artifacts and realizing strong cross-model generalization capability.
	
	Future work will focus on further improving the practical applicability of the method in complex scenarios.
	
	\bibliography{refs}

\begin{thebibliography}{10}

\bibitem{2014arXiv1406.2661G}
Ian~J. {Goodfellow}, Jean {Pouget-Abadie}, Mehdi {Mirza}, Bing {Xu}, David
  {Warde-Farley}, Sherjil {Ozair}, Aaron {Courville}, and Yoshua {Bengio}.
\newblock {Generative Adversarial Networks}.
\newblock {\em arXiv e-prints}, page arXiv:1406.2661, June 2014.

\bibitem{2013arXiv1312.6114K}
Diederik~P {Kingma} and Max {Welling}.
\newblock {Auto-Encoding Variational Bayes}.
\newblock {\em arXiv e-prints}, page arXiv:1312.6114, December 2013.

\bibitem{2019arXiv191211035W}
Sheng-Yu {Wang}, Oliver {Wang}, Richard {Zhang}, Andrew {Owens}, and Alexei~A.
  {Efros}.
\newblock {CNN-generated images are surprisingly easy to spot... for now}.
\newblock {\em arXiv e-prints}, page arXiv:1912.11035, December 2019.

\bibitem{Liu_2020_CVPR}
Zhengzhe Liu, Xiaojuan Qi, and Philip~H.S. Torr.
\newblock Global texture enhancement for fake face detection in the wild.
\newblock In {\em Proceedings of the IEEE/CVF Conference on Computer Vision and
  Pattern Recognition (CVPR)}, June 2020.

\bibitem{2023arXiv231210461T}
Chuangchuang {Tan}, Huan {Liu}, Yao {Zhao}, Shikui {Wei}, Guanghua {Gu}, Ping
  {Liu}, and Yunchao {Wei}.
\newblock {Rethinking the Up-Sampling Operations in CNN-based Generative
  Network for Generalizable Deepfake Detection}.
\newblock {\em arXiv e-prints}, page arXiv:2312.10461, December 2023.

\bibitem{2023arXiv231112397Z}
Nan {Zhong}, Yiran {Xu}, Sheng {Li}, Zhenxing {Qian}, and Xinpeng {Zhang}.
\newblock {PatchCraft: Exploring Texture Patch for Efficient AI-generated Image
  Detection}.
\newblock {\em arXiv e-prints}, page arXiv:2311.12397, November 2023.

\bibitem{10.5555/3524938.3525242}
Joel Frank, Thorsten Eisenhofer, Lea Sch\"{o}nherr, Asja Fischer, Dorothea
  Kolossa, and Thorsten Holz.
\newblock Leveraging frequency analysis for deep fake image recognition.
\newblock In {\em Proceedings of the 37th International Conference on Machine
  Learning}, ICML'20. JMLR.org, 2020.

\bibitem{2024arXiv240307240T}
Chuangchuang {Tan}, Yao {Zhao}, Shikui {Wei}, Guanghua {Gu}, Ping {Liu}, and
  Yunchao {Wei}.
\newblock {Frequency-Aware Deepfake Detection: Improving Generalizability
  through Frequency Space Learning}.
\newblock {\em arXiv e-prints}, page arXiv:2403.07240, March 2024.

\bibitem{Qin2020FcaNetFC}
Zequn Qin, Pengyi Zhang, Fei Wu, and Xi~Li.
\newblock Fcanet: Frequency channel attention networks.
\newblock {\em 2021 IEEE/CVF International Conference on Computer Vision
  (ICCV)}, pages 763--772, 2020.

\bibitem{2023arXiv230309295W}
Zhendong {Wang}, Jianmin {Bao}, Wengang {Zhou}, Weilun {Wang}, Hezhen {Hu},
  Hong {Chen}, and Houqiang {Li}.
\newblock {DIRE for Diffusion-Generated Image Detection}.
\newblock {\em arXiv e-prints}, page arXiv:2303.09295, March 2023.

\bibitem{2024arXiv240317465L}
Yunpeng {Luo}, Junlong {Du}, Ke~{Yan}, and Shouhong {Ding}.
\newblock {LaRE$^2$: Latent Reconstruction Error Based Method for
  Diffusion-Generated Image Detection}.
\newblock {\em arXiv e-prints}, page arXiv:2403.17465, March 2024.

\bibitem{2024arXiv241207140C}
Beilin {Chu}, Xuan {Xu}, Xin {Wang}, Yufei {Zhang}, Weike {You}, and Linna
  {Zhou}.
\newblock {FIRE: Robust Detection of Diffusion-Generated Images via
  Frequency-Guided Reconstruction Error}.
\newblock {\em arXiv e-prints}, page arXiv:2412.07140, December 2024.

\bibitem{10203908}
Chuangchuang Tan, Yao Zhao, Shikui Wei, Guanghua Gu, and Yunchao Wei.
\newblock Learning on gradients: Generalized artifacts representation for
  gan-generated images detection.
\newblock In {\em 2023 IEEE/CVF Conference on Computer Vision and Pattern
  Recognition (CVPR)}, pages 12105--12114, 2023.

\bibitem{10.1007/978-3-031-19781-9_6}
Bo~Liu, Fan Yang, Xiuli Bi, Bin Xiao, Weisheng Li, and Xinbo Gao.
\newblock Detecting generated images by real images.
\newblock In {\em Computer Vision – ECCV 2022: 17th European Conference, Tel
  Aviv, Israel, October 23–27, 2022, Proceedings, Part XIV}, page 95–110,
  Berlin, Heidelberg, 2022. Springer-Verlag.

\bibitem{Wang_2020_CVPR}
Sheng-Yu Wang, Oliver Wang, Richard Zhang, Andrew Owens, and Alexei~A. Efros.
\newblock Cnn-generated images are surprisingly easy to spot... for now.
\newblock In {\em Proceedings of the IEEE/CVF Conference on Computer Vision and
  Pattern Recognition (CVPR)}, June 2020.

\bibitem{pmlr-v119-frank20a}
Joel Frank, Thorsten Eisenhofer, Lea Sch{\"o}nherr, Asja Fischer, Dorothea
  Kolossa, and Thorsten Holz.
\newblock Leveraging frequency analysis for deep fake image recognition.
\newblock In Hal~Daumé III and Aarti Singh, editors, {\em Proceedings of the
  37th International Conference on Machine Learning}, volume 119 of {\em
  Proceedings of Machine Learning Research}, pages 3247--3258. PMLR, 13--18 Jul
  2020.

\bibitem{Tan2024FrequencyAwareDD}
Chuangchuang Tan, Yao Zhao, Shikui Wei, Guanghua Gu, Ping Liu, and Yunchao Wei.
\newblock Frequency-aware deepfake detection: Improving generalizability
  through frequency space domain learning.
\newblock In {\em AAAI Conference on Artificial Intelligence}, 2024.

\bibitem{2014arXiv1411.1784M}
Mehdi {Mirza} and Simon {Osindero}.
\newblock {Conditional Generative Adversarial Nets}.
\newblock {\em arXiv e-prints}, page arXiv:1411.1784, November 2014.

\bibitem{2017arXiv171010196K}
Tero {Karras}, Timo {Aila}, Samuli {Laine}, and Jaakko {Lehtinen}.
\newblock {Progressive Growing of GANs for Improved Quality, Stability, and
  Variation}.
\newblock {\em arXiv e-prints}, page arXiv:1710.10196, October 2017.

\bibitem{brock2018large}
Andrew Brock, Jeff Donahue, and Karen Simonyan.
\newblock Large scale {GAN} training for high fidelity natural image synthesis.
\newblock In {\em International Conference on Learning Representations}, 2019.

\bibitem{Choi_2018_CVPR}
Yunjey Choi, Minje Choi, Munyoung Kim, Jung-Woo Ha, Sunghun Kim, and Jaegul
  Choo.
\newblock Stargan: Unified generative adversarial networks for multi-domain
  image-to-image translation.
\newblock In {\em Proceedings of the IEEE Conference on Computer Vision and
  Pattern Recognition (CVPR)}, June 2018.

\bibitem{Karras_2019_CVPR}
Tero Karras, Samuli Laine, and Timo Aila.
\newblock A style-based generator architecture for generative adversarial
  networks.
\newblock In {\em Proceedings of the IEEE/CVF Conference on Computer Vision and
  Pattern Recognition (CVPR)}, June 2019.

\bibitem{pmlr-v37-sohl-dickstein15}
Jascha Sohl-Dickstein, Eric Weiss, Niru Maheswaranathan, and Surya Ganguli.
\newblock Deep unsupervised learning using nonequilibrium thermodynamics.
\newblock In Francis Bach and David Blei, editors, {\em Proceedings of the 32nd
  International Conference on Machine Learning}, volume~37 of {\em Proceedings
  of Machine Learning Research}, pages 2256--2265, Lille, France, 07--09 Jul
  2015. PMLR.

\bibitem{nichol2021improved}
Alexander~Quinn Nichol and Prafulla Dhariwal.
\newblock Improved denoising diffusion probabilistic models, 2021.

\bibitem{2022arXiv220900796Y}
Ling {Yang}, Zhilong {Zhang}, Yang {Song}, Shenda {Hong}, Runsheng {Xu}, Yue
  {Zhao}, Wentao {Zhang}, Bin {Cui}, and Ming-Hsuan {Yang}.
\newblock {Diffusion Models: A Comprehensive Survey of Methods and
  Applications}.
\newblock {\em arXiv e-prints}, page arXiv:2209.00796, September 2022.

\bibitem{2021arXiv211210741N}
Alex {Nichol}, Prafulla {Dhariwal}, Aditya {Ramesh}, Pranav {Shyam}, Pamela
  {Mishkin}, Bob {McGrew}, Ilya {Sutskever}, and Mark {Chen}.
\newblock {GLIDE: Towards Photorealistic Image Generation and Editing with
  Text-Guided Diffusion Models}.
\newblock {\em arXiv e-prints}, page arXiv:2112.10741, December 2021.

\bibitem{2021arXiv211114822G}
Shuyang {Gu}, Dong {Chen}, Jianmin {Bao}, Fang {Wen}, Bo~{Zhang}, Dongdong
  {Chen}, Lu~{Yuan}, and Baining {Guo}.
\newblock {Vector Quantized Diffusion Model for Text-to-Image Synthesis}.
\newblock {\em arXiv e-prints}, page arXiv:2111.14822, November 2021.

\bibitem{NEURIPS2021_49ad23d1}
Prafulla Dhariwal and Alexander Nichol.
\newblock Diffusion models beat gans on image synthesis.
\newblock In M.~Ranzato, A.~Beygelzimer, Y.~Dauphin, P.S. Liang, and J.~Wortman
  Vaughan, editors, {\em Advances in Neural Information Processing Systems},
  volume~34, pages 8780--8794. Curran Associates, Inc., 2021.

\bibitem{2023JFM...957A...6Y}
Mustafa~Z. {Yousif}, Meng {Zhang}, Linqi {Yu}, Ricardo {Vinuesa}, and HeeChang
  {Lim}.
\newblock {A transformer-based synthetic-inflow generator for spatially
  developing turbulent boundary layers}.
\newblock {\em Journal of Fluid Mechanics}, 957:A6, February 2023.

\bibitem{2023arXiv230210174O}
Utkarsh {Ojha}, Yuheng {Li}, and Yong~Jae {Lee}.
\newblock {Towards Universal Fake Image Detectors that Generalize Across
  Generative Models}.
\newblock {\em arXiv e-prints}, page arXiv:2302.10174, February 2023.

\bibitem{2024arXiv241204292H}
Zhenglin {Huang}, Jinwei {Hu}, Xiangtai {Li}, Yiwei {He}, Xingyu {Zhao}, Bei
  {Peng}, Baoyuan {Wu}, Xiaowei {Huang}, and Guangliang {Cheng}.
\newblock {SIDA: Social Media Image Deepfake Detection, Localization and
  Explanation with Large Multimodal Model}.
\newblock {\em arXiv e-prints}, page arXiv:2412.04292, December 2024.

\bibitem{10.1007/978-3-030-58574-7_7}
Lucy Chai, David Bau, Ser-Nam Lim, and Phillip Isola.
\newblock What makes fake images detectable? understanding properties that
  generalize.
\newblock In {\em Computer Vision – ECCV 2020: 16th European Conference,
  Glasgow, UK, August 23–28, 2020, Proceedings, Part XXVI}, page 103–120,
  Berlin, Heidelberg, 2020. Springer-Verlag.

\bibitem{He2015DeepRL}
Kaiming He, X.~Zhang, Shaoqing Ren, and Jian Sun.
\newblock Deep residual learning for image recognition.
\newblock {\em 2016 IEEE Conference on Computer Vision and Pattern Recognition
  (CVPR)}, pages 770--778, 2015.

\bibitem{pmlr-v97-tan19a}
Mingxing Tan and Quoc Le.
\newblock {E}fficient{N}et: Rethinking model scaling for convolutional neural
  networks.
\newblock In Kamalika Chaudhuri and Ruslan Salakhutdinov, editors, {\em
  Proceedings of the 36th International Conference on Machine Learning},
  volume~97 of {\em Proceedings of Machine Learning Research}, pages
  6105--6114. PMLR, 09--15 Jun 2019.

\bibitem{Fridrich2012RichMF}
Jessica~J. Fridrich and Jan Kodovsk{\'y}.
\newblock Rich models for steganalysis of digital images.
\newblock {\em IEEE Transactions on Information Forensics and Security},
  7:868--882, 2012.

\bibitem{2015arXiv150603365Y}
Fisher {Yu}, Ari {Seff}, Yinda {Zhang}, Shuran {Song}, Thomas {Funkhouser}, and
  Jianxiong {Xiao}.
\newblock {LSUN: Construction of a Large-scale Image Dataset using Deep
  Learning with Humans in the Loop}.
\newblock {\em arXiv e-prints}, page arXiv:1506.03365, June 2015.

\bibitem{2020arXiv200200133L}
Zhengzhe {Liu}, Xiaojuan {Qi}, and Philip {Torr}.
\newblock {Global Texture Enhancement for Fake Face Detection in the Wild}.
\newblock {\em arXiv e-prints}, page arXiv:2002.00133, January 2020.

\bibitem{Ju2022FusingGA}
Yan Ju, Shan Jia, Lipeng Ke, Hongfei Xue, Koki Nagano, and Siwei Lyu.
\newblock Fusing global and local features for generalized ai-synthesized image
  detection.
\newblock {\em 2022 IEEE International Conference on Image Processing (ICIP)},
  pages 3465--3469, 2022.

\bibitem{2008Visualizing}
Van Der~Maaten Laurens and Geoffrey Hinton.
\newblock Visualizing data using t-sne.
\newblock {\em Journal of Machine Learning Research}, 9(2605):2579--2605, 2008.

\end{thebibliography}
	\newpage 
	\clearpage 
	
	\appendix 
	\renewcommand{\appendixname}{Appendix} 
	\begin{table*}[htbp]
		\centering
		\small
		\caption{The convolution kernels for high-pass filters a-g in SRM (Spatial Rich Model) and their rotation directions}
		\label{tab:conv_kernels}
		\resizebox{0.675\textwidth}{!}{
			\begin{tabular}{l c l}
				\toprule
				num & kernel & rotation directions \\
				\midrule
				a & 
				$\begin{pmatrix}
					0 & 0 & 0 & 0 & 0 \\
					0 & 1 & 0 & 0 & 0 \\
					0 & -1 & 0 & 0 & 0 \\
					0 & 0 & 0 & 0 & 0 \\
					0 & 0 & 0 & 0 & 0
				\end{pmatrix}$ & 
				$\nearrow, \rightarrow, \searrow, \downarrow, \swarrow, \leftarrow, \nwarrow, \uparrow$ \\
				\addlinespace[0.3em]
				
				b & 
				$\begin{pmatrix}
					0 & 0 & -1 & 0 & 0 \\
					0 & 0 & 3 & 0 & 0 \\
					0 & 0 & -3 & 0 & 0 \\
					0 & 0 & 1 & 0 & 0 \\
					0 & 0 & 0 & 0 & 0
				\end{pmatrix}$ & 
				$\nearrow, \rightarrow, \searrow, \downarrow, \swarrow, \leftarrow, \nwarrow, \uparrow$ \\
				\addlinespace[0.3em]
				
				c & 
				$\begin{pmatrix}
					0 & 0 & 0 & 0 & 0 \\
					0 & 1 & 0 & 0 & 0 \\
					0 & 0 & -2 & 0 & 0 \\
					0 & 0 & 1 & 0 & 0 \\
					0 & 0 & 0 & 0 & 0
				\end{pmatrix}$ & 
				$\rightarrow, \downarrow, \nearrow, \searrow$ \\
				\addlinespace[0.3em]
				
				d & 
				$\begin{pmatrix}
					0 & 0 & 0 & 0 & 0 \\
					0 & -1 & 2 & -1 &0 \\
					0 & 2 & -4 & 2 & 0 \\
					0 & 0 & 0 & 0 & 0 \\ 
					0 & 0 & 0 & 0 & 0
				\end{pmatrix}$ & 
				$\rightarrow, \downarrow, \leftarrow, \uparrow$ \\
				\addlinespace[0.3em]
				
				e & 
				$\begin{pmatrix}
					-1 & 2 & -2 & 2 & -1 \\
					2 & -6 & 8 & -6 & 2 \\
					-2 & 8 & -12 & 8 & -2 \\
					0 & 0 & 0 & 0 & 0 \\
					0 & 0 & 0 & 0 & 0
				\end{pmatrix}$ & 
				$\rightarrow, \downarrow, \leftarrow$ \\
				\addlinespace[0.3em]
				
				f & 
				$\begin{pmatrix}
					0 & 0 & 0 & 0 & 0 \\
					0 & -1 & 2 & -1 & 0 \\
					0 & 2 & -4 & 2 & 0 \\
					0 & -1 & 2 & -1 & 0 \\
					0 & 0 & 0 & 0 & 0
				\end{pmatrix}$ & None \\
				\addlinespace[0.3em]
				
				g & 
				$\begin{pmatrix}
					-1 & 2 & -2 & 2 & -1 \\
					2 & -6 & 8 & -6 & 2 \\
					-2 & 8 & -12 & 8 & -2 \\
					2 & -6 & 8 & -6 & 2 \\
					-1 & 2 & -2 & 2 & -1
				\end{pmatrix}$ & None \\
				
				\bottomrule
			\end{tabular}
		}
	\end{table*}
	
	\begin{table*}[!t]  
		\centering
		\caption{Design of Cascaded Discriminator Module} 
		\label{tab:discriminator_design}
		\begin{tabular*}{\textwidth}{@{\extracolsep{\fill}} l c c c @{}}
			\toprule
			Type                  & Input Channels & Normalization & Activation Function \\
			\midrule
			Convolutional Layer   & 64             & BN            & ReLU                \\
			Convolutional Layer   & 32             & BN            & ReLU                \\
			Convolutional Layer   & 32             & BN            & ReLU                \\
			Convolutional Layer   & 32             & BN            & ReLU                \\
			Average Pooling Layer & 32             & --            & --                  \\
			Convolutional Layer   & 32             & BN            & ReLU                \\
			Convolutional Layer   & 32             & BN            & ReLU                \\
			Average Pooling Layer & 32             & --            & --                  \\
			Convolutional Layer   & 32             & BN            & ReLU                \\
			Convolutional Layer   & 32             & BN            & ReLU                \\
			Average Pooling Layer & 32             & --            & --                  \\
			Convolutional Layer   & 32             & BN            & ReLU                \\
			Convolutional Layer   & 32             & BN            & ReLU                \\
			Adaptive Avg Pooling Layer & 32       & --            & --                  \\
			Fully Connected Layer & --             & --            & --                  \\
			\bottomrule
		\end{tabular*}
	\end{table*}
	
\end{document}